\documentclass[sigconf]{acmart}
\AtBeginDocument{%
  }

\usepackage{pifont}
\DeclareMathOperator*{\minimize}{minimize}
\usepackage{subcaption}
\usepackage{enumitem}
\usepackage{multirow}
\usepackage[svgnames]{xcolor} 
\usepackage{makecell}
\usepackage[table]{xcolor}

\newcommand{\proposed}{GLASS}
\newcommand{\stdtiny}[1]{{\small $\pm$ #1}}

\copyrightyear{2026}
\acmYear{2026}
\setcopyright{cc}
\setcctype{by}
\acmConference[KDD '26]{Proceedings of the 32nd ACM SIGKDD Conference on Knowledge Discovery and Data Mining V.2}{August 09--13, 2026}{Jeju Island, Republic of Korea}
\acmBooktitle{Proceedings of the 32nd ACM SIGKDD Conference on Knowledge Discovery and Data Mining V.2 (KDD '26), August 09--13, 2026, Jeju Island, Republic of Korea}
\acmDOI{10.1145/3770855.3817677}
\acmISBN{979-8-4007-2259-2/2026/08}

\begin{document}

\title{
Explaining Black-Box Language Models: Learning to Optimize Linguistically-Structured Word Subsets
}
\author{Minyoung Hwang}
\authornote{Both authors contributed equally to this research.}
\affiliation{%
  \institution{Korea University}
  \country{Republic of Korea}
}
\email{minyoung58@korea.ac.kr}

\author{Seokhyun Lee}
\authornotemark[1]
\affiliation{%
  \institution{Korea University}
  \country{Republic of Korea}
}
\email{nshuhsn@korea.ac.kr}

\author{Changhee Lee}
\affiliation{%
  \institution{Korea University}
  \country{Republic of Korea}
}
\email{changheelee@korea.ac.kr}

\newcommand{\circlegreen}{\tikz\draw[draw=black,fill=green] (0,0) circle (5pt);}
\newcommand{\circleyellow}{\tikz\draw[draw=black,fill=yellow] (0,0) circle (5pt);}
\newcommand{\circlered}{\tikz\draw[draw=black,fill=red] (0,0) circle (5pt);}
\newcommand{\circlegray}{\tikz\draw[fill=gray!40,draw=black] (0,0) circle (5pt);}

\newcommand{\cmark}{\textcolor{black}{\scalebox{1.3}{\ding{51}}}}  %
\newcommand{\xmark}{\textcolor{red}{\scalebox{1.3}{\ding{55}}}}    %
\newcommand{\nmark}{\textcolor{black}{\scalebox{1.5}{\ding{84}}}}  %
\newcommand{\trianglmark}{\textcolor{orange}{\large$\triangle$}}
\newcommand{\subentry}{\hspace{1em}\textemdash~}

\newcommand{\defeq}{\overset{\text{\tiny def}}{=}}
\newcommand{\ldefeq}{\mathrel{\raisebox{-0.3ex}{$\defeq$}}}

\newcommand{\hrationale}[1]{{\setlength{\fboxsep}{1pt}\colorbox{LightGreen}{#1}}}
\newcommand{\hmissed}[1]{{\setlength{\fboxsep}{1pt}\colorbox{LightCoral}{#1}}}
\newcommand{\hextra}[1]{{\setlength{\fboxsep}{1pt}\colorbox{Orange}{#1}}}

\begin{abstract}
As deep language models (DLMs) are increasingly deployed in high-stakes domains such as healthcare, understanding their decision rationale becomes paramount for ensuring trust, safety, and accountability. 
However, achieving this vital level of interpretability is particularly challenging when these DLMs operate as \textit{black-box} systems (e.g., via APIs), where access to internal model states (e.g., parameters, gradients) is restricted. Despite numerous efforts, existing explanation methods often fail to concurrently satisfy three key desiderata: (i) \textit{inference-time efficiency}, (ii) \textit{black-box compatibility} without inducing out-of-distribution behavior, and (iii) \textit{comprehensible explanations} grounded in the input's linguistic structure. To address these challenges, we propose a method that explains predictions of DLMs by selecting a small, informative subset of input words. We formulate this as an amortized optimization problem, enabling efficient one-shot inference without the need for input-specific search. Our selection policy is trained via REINFORCE-style policy gradients, allowing discrete word selection in a fully gradient-free setting. To enhance interpretability and align with human linguistic intuition, we integrate graph-structured knowledge into this selection process, fostering linguistically coherent subsets that result in explanations both highly informative and cognitively meaningful to end-users. We evaluated our method on diverse DLM architectures and multiple real-world datasets. It consistently identifies word subsets with enhanced discriminative power and stronger alignment with linguistically salient cues, outperforming both conventional black-box compatible methods and gradient-based approaches that are given oracle access to the black-box model's gradients for a more challenging benchmark. Our code is available at 
\href{https://github.com/ggomaeng514/GLASS}{\underline{here}}.
\end{abstract}




\maketitle

\renewcommand{\shortauthors}{Minyoung Hwang, Seokhyun Lee, and Changhee Lee}

\section{Introduction}
Deep Language models (DLMs) have become foundational to diverse applications ranging from clinical decision support and legal analysis to scientific discovery \cite{lee_biobert_Bioinformatics2020, chalkidis_legalbert_EMNLP2020, beltagy_scibert_EMNLP2019}. As these models are increasingly deployed in high-stakes settings, the demand for understanding model predictions has become increasingly critical. 
This need is severely compounded when DLMs are provided as \textbf{\textit{strict black-box}} systems (e.g., via APIs), which typically restrict access to internal parameters, gradients, and may even obscure crucial details of their input processing pipelines, including tokenization schemes.
When users are unable to understand the rationale behind a model's decision, the system remains a powerful yet opaque tool, raising concerns around trust, safety, and accountability.

To enable widespread adoption of model explanations in real-world workflows --- especially, those operating at scale or under tight latency constraints --- fast inference is crucial.
Memoryless explanation methods \cite{sundararajan_ig_ICML2017, binder_LRP_ICANN2016, shrikumar_DeepLIFT_ICML2017, zeiler_visualizing_ECCV2014, lundberg_shap_NeurIPS2017, ribeiro_LIME_KDD2016} often fall short in this regard, as they require multiple queries to the model or costly per-instance optimization to identify salient features, making them impractical for time-sensitive environments. Amortized explanation methods \cite{barkan_LTX_ICDM2023, yoon_invase_ICLR2018, chen_l2x_ICML2018, schwab_cxplain_NeurIPS2019} present a compelling alternative by training a reusable explainer model capable of generating explanations in a single pass that generalize across inputs. 
However, even though these approaches avoid direct reliance on model gradients, applying them to large-scale DLMs remains challenging --- training a sufficiently expressive surrogate often requires substantial data and computational resources, making them impractical for many real-world scenarios.
While both memoryless and amortized methods provide local (instance-specific) explanations, they differ fundamentally in how these explanations are produced.
A summary comparison is presented in Table \ref{table:summary}. (See Section \ref{section:related_work} for more details).

\begin{table}[!t]

\centering

\caption{\small A comparative summary of explanation methods.}  \vspace{-2.5mm}
\label{table:summary}
\renewcommand{\arraystretch}{1.0}
\resizebox{0.48\textwidth}{!}{
    \begin{tabular}{lcccccc}
    \toprule
    \multirow{2}{*}{\textbf{Method}} &  \textbf{Black-box}&\textbf{Discrete}& \textbf{Linguistic}&\textbf{Black-box}&  \multicolumn{2}{c}{\textbf{Cost}} \\
    &\textbf{Compatibility} &\textbf{Space} &\textbf{Structure} &\textbf{Reuse} &\textbf{Training} &\textbf{Inference} \\
    \midrule
    \textit{Memoryless}&  && &&  &\\
     \hspace{0.3em}
    IG
    \cite{sundararajan_ig_ICML2017}& \xmark& \xmark& \xmark& \cmark & None&High\\
     \hspace{0.3em}
    LRP, DeepLIFT
    \cite{binder_LRP_ICANN2016, shrikumar_DeepLIFT_ICML2017}& \xmark & \xmark& \xmark& \cmark & None&Low \\
     \hspace{0.3em} 
    PDA, KernelSHAP 
    \cite{zintgraf_featureocclusion_ICLR2019, lundberg_shap_NeurIPS2017}&  \cmark &\cmark & \xmark&\cmark & None&High\\
     \hspace{0.3em} LIME \cite{ribeiro_LIME_KDD2016}&  \cmark &\cmark & \xmark&\xmark & None&High \\
    \midrule
    \textit{Amortized}&         && &&                &\\
     \hspace{0.3em} LTX \cite{barkan_LTX_ICDM2023}& \xmark& \xmark & \xmark& \cmark & Low&Low\\
     \hspace{0.3em} INVASE \cite{yoon_invase_ICLR2018}& \cmark & \cmark & \xmark& \xmark & Moderate&Low\\
    \hspace{0.3em} L2X \cite{chen_l2x_ICML2018}&  \cmark &\xmark & \xmark&\xmark &  Moderate&Low\\
    \hspace{0.3em} CXPlain \cite{schwab_cxplain_NeurIPS2019}&  \cmark &\cmark & \xmark&\xmark &  High&Low\\
    \midrule
    \rowcolor[HTML]{FDEBD0}\textbf{Ours}&  \cmark &\cmark & \cmark &\cmark &  Low&Low\\
    \bottomrule 
    \end{tabular} 
}  
\vspace{-4mm}
\end{table} 

Compounding the issue, \textit{comprehensibility} of explanations is often overlooked in many existing explanation methods. Previous methods concentrate primarily on identifying input tokens that are most influential to the model's prediction, without adequately considering whether the selected subset forms a coherent, meaningful span from a linguistic perspective. For instance, an explanation composed of scattered or grammatically disconnected words might faithfully reflect certain aspects of the model's internal decision process, yet fail to help a human understand the underlying rationale or trust the decision \cite{Matthew_chunk_PNAS2017}. In such cases, technical fidelity alone is insufficient. 
As DLMs are increasingly integrated into human-centered applications --- often under strict black-box settings --- it becomes critical to generate explanations that align with human linguistic expectations, which is key to ensuring interpretability, usability, and responsible deployment.

\vspace{1mm}
\noindent
\textbf{Contributions.}~~ We propose a novel method that jointly satisfies the three core desiderata for explaining strict black-box DLMs: (i) \textbf{\textit{efficiency}}, (ii) \textbf{\textit{black-box compatibility}}, and (iii) \textbf{\textit{comprehensible explanations}}. 
We cast the explanation task as an \textit{amortized optimization} problem, training a reusable selector that enables fast, single-pass inference at test time. 
Our selector is trained using \textit{policy gradients} --- defining a Bernoulli policy for selecting input words and using rewards from the black-box model's outputs --- which ensures compatibility with strictly opaque systems and avoids the need for computationally expensive and potentially unfaithful surrogate models (specifically, for large DLMs).
Furthermore, to enhance the comprehensibility of the resulting explanations, we incorporate graph-structured linguistic knowledge (e.g., syntactic dependency graphs) into the selection mechanism. 
Unlike methods that view text as a simple sequence of tokens, this structural guidance biases our selector towards identifying subsets that are both predictive of the model’s output and linguistically coherent, thereby providing rationales that are more readily understood by human users.

\begin{figure*}[!t]
    \centering
    \setlength{\abovecaptionskip}{0.1mm}
    \includegraphics[trim={0cm 0cm 0cm 0cm}, clip, width=0.85 \linewidth]
    {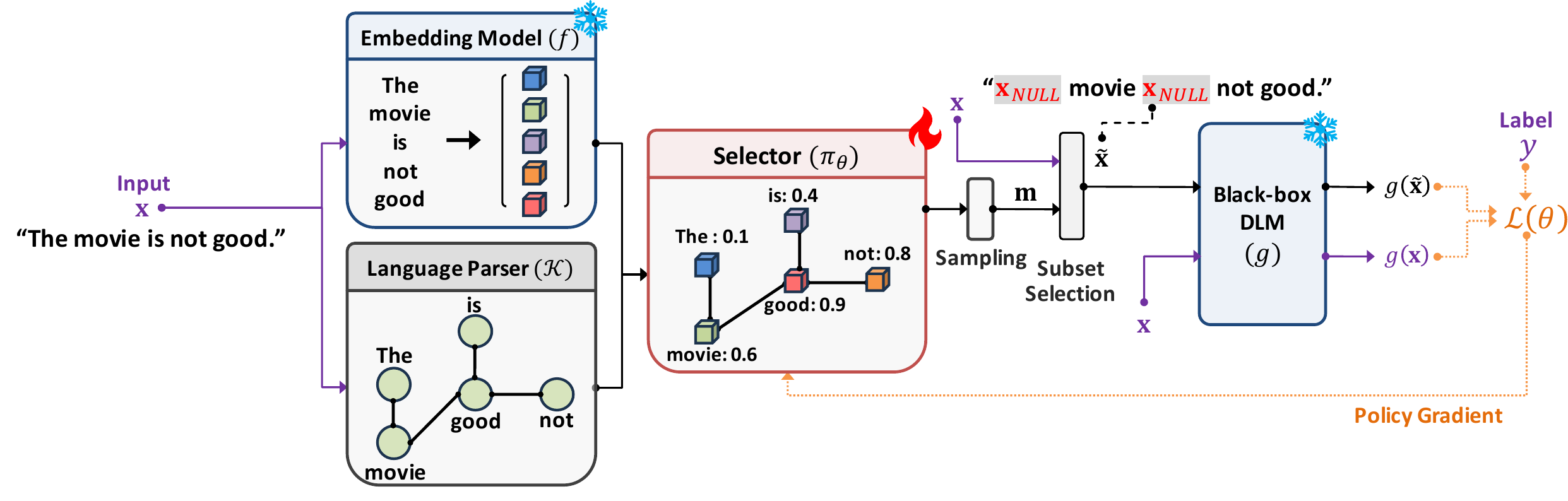} 
    \caption{\small A schematic illustration of \proposed. The selector, $\pi_\theta$, is trained via policy gradient, while the pre-trained word embedding model, $f$, and the black-box DLM, $g$, remain fixed. A language parser (e.g., SpaCy’s dependency parser \cite{honnibal_spacy_2020}) provides a graph representing the input's linguistic structure to inform the selector.}
    \label{fig:block_diagram}
    \Description{model figure}
\vspace{-3.5mm}
\end{figure*}

\section{Problem Formulation} \label{sec:proble_formulation}

Suppose that we are given a DLM, $g: \mathcal{X}^T \rightarrow [0,1]^C$, trained to predict a target label $y \in \{1,\dots, C\}$ based on a variable-length input sequence $\mathbf {x}=(x_1,x_2,\dots, x_T) \in \mathcal{X}^T$ consisting of $T$ words, where $\mathcal{X}$ denotes the input (word) space. Note that the sequence length $T$ varies as the input sequence changes. 
In this work, we focus on the classification setting where the DLM has been fine-tuned or trained for a specific task; we also demonstrate its extensibility to generation tasks in Section \ref{main: summarization} and discuss the formulation for regression in Appendix \ref{appendix: regression}. 
We assume that $g$ is a \textit{strict black-box} function that we can evaluate the model output $g(\mathbf{x})$ for any given instance $\mathbf{x}$, but we do not have any knowledge of the internal model states such as parameters, gradients, and existence of particular tokens (e.g., \texttt{[MASK]},\texttt{[PAD]}).

Our \textit{goal of explanation} is to identify a subset of words in $\mathbf{x}$ that are most influential in determining the model output $g(\mathbf{x})$ given an input $\mathbf{x}$. To formalize this, we introduce a binary gating vector $\mathbf{m} = (m_1,m_2 \dots,m_T) \in \{0,1\}^{T}$, where each element $m_t$ specifies whether the $t$-th word in the input sequence is selected (i.e., $m_t = 1$) or not (i.e., $m_t = 0$). 
Then, the input sequence corrupted by $\mathbf{m}$ can be written as
\begin{equation} \label{eq:corrupted_input}
    \tilde{\mathbf{x}} = \mathbf{m}\odot \mathbf{x} + (\mathbf{1} - \mathbf{m}) \cdot x_{\texttt{NULL}}
\end{equation}
where $\odot$ is element-wise multiplication and $x_{\texttt{NULL}}$ is a word-level placeholder to fill the positions of not chosen words.
This formulation relies on the assumption that replacing unselected words with placeholders does not substantially alter the model's behavior.

\vspace{1mm}
\noindent
\textbf{Memoryless Optimization.}~~
For a given input sequence $\mathbf{x}$, identifying the important words can be achieved by selecting the set of words that make similar outputs when compared to those given the original input. This can be formally defined through the following optimization problem:
\begin{equation} \label{eq:memoryless_optimization}
    \minimize_{\mathbf{m} \in \{0,1\}^T} ~~ \ell \big( g(\mathbf{x}),\, g(\tilde{\mathbf{x}}) \big) \quad\quad  \text{subject to} ~~\|\mathbf{m}\|_0 < \delta
\end{equation}
where $\delta$ constrains the number of important words and $\ell$ is a loss function --- e.g., the KL divergence --- that quantifies the difference between the black-box model output based on the original input $\mathbf{x}$ and that based on the corrupted input $\tilde{\mathbf{x}}$. 
The optimal subset of words can be identified by evaluating how the model's output changes when the corresponding words are corrupted. However, this optimization is inherently \textit{memoryless}; it requires recomputing the solution independently for each input sequence, without exploiting information from previously processed (and potentially similar) sequences.
Furthermore, since the search space over word subsets is discrete, the computational cost of solving the optimization grows exponentially with the sequence length.

\vspace{1mm}

\noindent
\textbf{Amortized Optimization.}~~
To address above limitations, we amortize the cost of solving the per-instance optimization problems in \eqref{eq:memoryless_optimization} over a population of input sequences by introducing a selection mechanism parameterized by a neural network.
Specifically, we define a selector (parameterized by $\theta$), $\pi_{\theta}: \mathcal{X}^T \rightarrow [0,1]^T$, that stochastically maps an input sequence to a binary gating vector, i.e., $\mathbf{m} \sim \texttt{Bern}(\pi_\theta(\mathbf{x}))$. 
Then, we can reformulate \eqref{eq:memoryless_optimization} as:
\begin{equation} \label{eq:amortized_optimization} \vspace{-0.5mm}
    \minimize_{\theta} \ \mathbb{E}_{\mathbf{x}\sim p_X} \mathbb{E}_{\mathbf{m} \sim \texttt{Bern}(\pi_\theta(\mathbf{x}))}\big[\ell \big(g(\mathbf{x}) \ ,\  g(\tilde{\mathbf{x}})\big) + \lambda||\mathbf{m}||_0\big] 
\end{equation}
where $\lambda \geq 0$ balances fidelity and sparsity by controlling the number of words selected. Hereafter, we will denote this objective as $\mathcal{L}(\theta)$. 
This amortized formulation is particularly appealing because, in many practical settings, there exists substantial structure and correlation among the optimal gating vectors across different input sequences, allowing us to replace the combinatorial search in \eqref{eq:memoryless_optimization} with a more tractable optimization over continuous parameters $\theta$. 
More importantly, we deviate from estimating the black-box function to approximate $g(\tilde{\mathbf{x}})$ --- a common step in previous joint amortized explanation approaches \cite{yoon_invase_ICLR2018, chen_l2x_ICML2018, jethani_REAL-X_AISTATS2021, panda_ICFS_CVPR2021} --- as training a sufficiently complex predictor to match the behavior of large-scale DLMs typically requires considerable data and computational overhead, ultimately limiting the practical utility of such approaches.

Unfortunately, identifying important word subsets via \eqref{eq:amortized_optimization} to explain black-box DLMs poses following challenges:  (i) \textit{Gradients are inaccessible}: Many black-box DLMs (particularly those accessed via APIs) do not provide access to the internal gradients, which makes learning to select gate vectors using backpropagation from model outputs inapplicable. (ii) \textit{Meaning in language emerges from compositional structure}: DLMs rely heavily on syntactic and semantic interactions between words to make predictions. Methods failing to account for such dependencies risk producing explanations that misrepresent the model's actual decision-making process.

\section{Method}

To address the challenges outlined above, we introduce a novel method designed to produce faithful and interpretable explanations for black-box DLMs, which we refer to as \textbf{G}raph-aware, \textbf{L}abel-aligned, \textbf{A}mortized \textbf{S}ubset \textbf{S}election (\textbf{GLASS}). Our approach combines (i) a \textit{policy gradient-based learning strategy} to address the discrete and non-differentiable nature of black-box DLMs, and (ii) \textit{inductive bias} by incorporating the input's structural knowledge to encourage the selection of linguistically coherent word subsets. %

\subsection{Optimizing the Selector via Policy Gradients} \label{sec: Policy Gradient}

In many applications, explanations need to be anchored to ground-truth outcomes, such as identifying a minimal subset of input words that leads the black-box model to the correct classification. 
Thus, we employ a supervised loss term for $\ell(g(\mathbf{x}), g(\tilde{\mathbf{x}}))$ in \eqref{eq:amortized_optimization} by incorporating the ground-truth label $y$ to guide the selection process: 
\begin{equation} \label{eq:loss_CE}
        \ell_y \big(g(\mathbf{x}),g(\tilde{\mathbf{x}}) \big) = \sum_{c=1}^{C} \mathbb{I}(y = c) \big(  \log (g(\mathbf{x})_c) - \log(g(\tilde{\mathbf{x}})_c) \big).
\end{equation}
Here, $g(\mathbf{x})_c$ and $g(\tilde{\mathbf{x}})_c$ denote the model's prediction for class $c$ given the entire input sequence $\mathbf{x}$ and the word subset $\tilde{\mathbf{x}}$, respectively.
The loss in \eqref{eq:loss_CE} can be interpreted as the difference between two cross-entropy terms: the first (constant with respect to $\theta$) term reflects the cross-entropy between the true label and the model's original prediction, while the second measures the cross-entropy based on the selected word subset. 
This formulation is designed for \textit{label-centric} explanation (as similarly proposed in \cite{ying_gnnexplainer_Neurips2019}) and naturally performs as baseline variance reduction, where the prediction on the full input serves as a fixed reference point, thereby stabilizing the gradient signal and reducing variance while minimizing the difference during training.

Now, we optimize the selector based on a REINFORCE-style policy gradient method \cite{williams_REINFORCE_ML1992}. 
This approach operates directly on the discrete action space of word selection, thereby obviating the need for continuous relaxations \cite{maddison_concrete_ICLR2017, liang_bayesian_JASA2018}. Such adherence to discrete inputs ensures that selected words are valid vocabulary items and mitigates the risk of potential out-of-distribution representations. 
Furthermore, this enables the selector to learn solely from sampled model outputs and corresponding scalar rewards, making it ideally suited for black-box DLMs where internal gradients are unavailable.
Formally, we can compute the gradient of our objective $\mathcal{L}(\theta)$ in \eqref{eq:amortized_optimization} with respect to the selector parameters $\theta$ as follows: \vspace{-0.1mm}
\begin{align} \label{eq:Reinforce_loss}
    \nabla_\theta \mathcal{L}(\theta) = & \mathbb{E}_{\mathbf{x}, y\sim p_{XY}} \Big[ \mathbb{E}_{\mathbf{m} \sim \texttt{Bern}(\pi_{\theta}(\mathbf{x}))} \Big[ \\
    &\quad\quad\nabla_\theta \log p_{\theta}(\mathbf{m}) \nonumber
    \cdot \big( \ell_y\big(g(\mathbf{x}),g(\tilde{\mathbf{x}}) \big) + \lambda||\mathbf{m}||_0 \big) \Big] \Big], \vspace{-0.1mm}
\end{align}
where $p_{\theta}(\mathbf{m}) = \prod_{t=1}^{T} \pi_{\theta}(\mathbf{x})_t^{m_t} (1 - \pi_{\theta}(\mathbf{x})_t)^{(1-m_t)}$ is the probability of generating the gate vector $\mathbf{m}$ given input $\mathbf{x}$, and is modeled as a multivariate Bernoulli distribution. Here, each $\pi_\theta(\mathbf{x})_t$ denotes the selection probability of the $t$-th word. See Appendix \ref{Reinforce_loss_derivation} for details.

\subsection{Incorporating Human Linguistic Intuition} \label{subsection:knowledge}
To achieve explanations that are interpretable and structurally grounded, we design our selector to leverage external \textit{knowledge} reflecting \textit{human linguistic intuition}, denoted as $\mathcal{K}(\mathbf{x})$, during the word selection process for any given input sequence $\mathbf{x}$. This knowledge is presented in the form of graph structures that explicitly define linguistic relationships within the input, which represent grammatical organization (e.g., syntactic dependency graphs), semantic connections (e.g., semantic relation networks), or other task-specific constraints.

A straightforward way to integrate linguistic structure is by applying graph Laplacian regularization to the selector's output, thereby encouraging the assignment of similar selection probabilities to structurally connected words. Although effective in smoothing the selector output, it relies on the strong assumption that edge connectivity directly translates to functional similarity.
However, as noted by \cite{kipf_gcn_ICLR2017}, graph edges often encode diverse relationships, such as contrastive, hierarchical, or task-specific dependencies, where such similarity is not guaranteed or even desired.  Please refer to Appendix \ref{appendix: Laplacian} for further details on how graph Laplacian regularization can be considered within our method.

Instead, we directly condition our selector on both input $\mathbf{x}$ and external structural knowledge $\mathcal{K}(\mathbf{x})$, allowing for dynamic adaptation to the specific relationships encoded in the input's linguistic structure. 
To achieve this, we implement the selector $\pi_\theta$ as a Graph Neural Network (GNN) that computes word-level selection probabilities by representing words in $\mathbf{x}$ as nodes and utilizing $\mathcal{K}(\mathbf{x})$ to define the graph edges.
First, we derive initial node representations for each word from contextualized embeddings generated by a pretrained DLM, $f: \mathcal{X}^{T} \rightarrow \mathbb{R}^{T\times d}$, where $\mathbb{R}^d$ denotes the embedding space. 
The embedding model $f$ is a separate component from the target black-box model $g$ and is used exclusively to provide input features for the selector, which is fixed during training. 
This allows us to leverage $f$'s robust representational power to capture rich relationships among input words, ensuring efficient feature extraction without the complexity of additional training. 
Then, we compute the selector output using the GNN architecture as follows:
\begin{equation} \label{eq:selector_notation}
    \quad  \pi_\theta(\mathbf{x}, \mathcal{K} (\mathbf{x}))_t = \sigma(h_t) \quad \text{for}~~t=1,\dots,T
\end{equation}
Here, $\mathbf{h} = (h_1, \dots, h_T) = \texttt{GNN}(f(\mathbf{x}), \mathcal{K}(\mathbf{x}))$ and $\sigma$ is a sigmoid function.
Note that multiple types of linguistic structure can be derived from a given input sequence \cite{vashishth_SynSemGCN_ACL2018}. 
we provide an extension of GLASS that uses multiple forms of external knowledge in Appendix \ref{appendix: extension_knowledge}.

\subsection{Sampling Important Word Subsets}

To adhere to strict black-box constraints --- where access to the model’s input processing pipeline, including tokenization, embedding mappings, or special token IDs (e.g., \texttt{[MASK]}, \texttt{[PAD]}), is entirely restricted --- we apply a word-level gating mechanism. Unselected words in each input sequence are replaced with a semantically neutral placeholder (e.g., ``the'' or an empty string) that is unlikely to affect the model's prediction. This substitution preserves both the input length and syntactic structure, avoiding unnatural artifacts due to word deletion or embedding-level interventions.

Finally, given a designated placeholder $x_{\text{NULL}}$, we construct a corrupted input sequence $\tilde{\mathbf{x}}$ from an original input $\mathbf{x}$ according to \eqref{eq:corrupted_input}. This involves sampling a binary gate vector $\mathbf{m} \sim \texttt{Bern}(\pi_\theta(\mathbf{x}, \mathcal{K}(\mathbf{x})))$, where the selector $\pi_\theta$ is conditioned on the original input $\mathbf{x}$ and external linguistic structure $\mathcal{K}(\mathbf{x})$. 
Unlike approaches that manipulate token embeddings or internal attention mechanisms, our word-level gating mechanism requires no assumptions about the target model’s vocabulary, input processing pipeline, or tokenizer behavior. This design ensures full compatibility with strict black-box settings. %

\section{Related Work} \label{section:related_work}

\textbf{Memoryless Explanation Methods.}~~ 
Perturbation-based methods estimate feature importance by observing changes in the output of the target model when individual features (e.g., a leave-one-feature-out method \cite{zintgraf_featureocclusion_ICLR2019}) or combinations of features (e.g., KernelSHAP \cite{lundberg_shap_NeurIPS2017}) are perturbed. 
Since these methods rely solely on the interactions between input and model output, they can offer model-agnostic explanations --- hence, applicable to strict black-box models --- without training an additional model to generate explanations. 
However, in practice, applying these methods to DLMs often necessitates implicit or explicit access to the model's input processing pipeline. This can involve managing special token IDs (e.g., \texttt{[MASK]}, \texttt{[PAD]}), or even directly manipulating token embeddings (e.g., zeroing out embeddings of unselected words), to ensure these words are effectively neutralized, thereby avoiding unintended model behavior caused by improper word corruption \cite{sanyal_discretizedIG_EMNLP2021}.

Backpropagation-based methods compute feature attributions in a single forward or backward pass through the target model. Integrated Gradients (IG)~\cite{sundararajan_ig_ICML2017} and its recent adaptation for discrete inputs in language models \cite{sanyal_discretizedIG_EMNLP2021, enguehard_sequentialIG_ACL2023} calculate feature importance by integrating the gradients of the target model along a path from a baseline to the input. 
Layer-wise Relevance Propagation \cite{binder_LRP_ICANN2016} and DeepLIFT \cite{shrikumar_DeepLIFT_ICML2017} instead redistribute the model's output to the input %
by tracing gradients or activation differences across layers. 

Overall, while training-free, many memoryless explanation methods are not readily applicable to \textit{strict black-box} DLMs because they rely on a degree of internal model access, such as special token IDs, token embeddings, gradient, or even layer-specific computations.

\vspace{1mm}
\noindent
\textbf{Amortization Explanation Methods.}~~
To address the computational inefficiency of memoryless methods during inference (due to multiple model evaluations for each input), many recent approaches have adopted a ``learning to explain'' framework. These methods train a reusable explainer (or selector) that generalizes across inputs by finding explanation patterns via a separate, trainable model. 

Joint amortized explanation methods \cite{yoon_invase_ICLR2018, chen_l2x_ICML2018, jethani_REAL-X_AISTATS2021, panda_ICFS_CVPR2021} jointly train two key components: a \textit{selector} to identify relevant input feature subsets, and a \textit{predictor} to estimate the target model's output distribution conditioned on these subsets. The overarching goal is to find feature subsets maximally predictive of the target model's output, often achieved by maximizing mutual information between the selected features and target outputs, or, equivalently, by minimizing the Kullback-Leibler (KL) divergence between the target model's output distribution and the distribution predicted using only the selected feature subset. 
However, L2X  \cite{chen_l2x_ICML2018} and its variants \cite{jethani_REAL-X_AISTATS2021, panda_ICFS_CVPR2021} rely on continuous relaxation to approximate discrete subset selection, which poses a fundamental challenge in natural language tasks, where soft-weighted combinations of word (token) embeddings do not correspond to any valid word. 
Such representations can induce out-of-distribution behavior \cite{hase_OOD_NeurIPS2021}, potentially compromising the fidelity and stability of the resulting explanations.

In contrast, surrogate explanation methods such as L2E~\cite{situ_L2E_ACL2021} and CXPlain~\cite{schwab_cxplain_NeurIPS2019} aim to approximate existing explanation strategies by learning from the input-output behavior of the target model. L2E distills instance-level explanations --- typically, generated by memoryless methods --- into a stable, fast explainer through supervised learning. Similarly, CXPlain trains a surrogate model using feature ablation scores as the ground truth. While efficient at inference, these methods require extensive training queries to construct supervised labels, making them computationally expensive upfront.

Alternatively, recent approaches like GELPE~\cite{agiollo_GELPE_AAMAS2024} create a global surrogate model to resolve inconsistencies among local explainers. In contrast, our work focuses on improving the intrinsic quality and linguistic coherence of the individual local explanation itself.

\vspace{1mm}
\noindent
\textbf{Graph-based Linguistic Structure in NLP.}~~
Linguistic graphs, such as syntactic dependency trees, are central to improving both performance and interpretability of NLP models. By encoding structural relationships beyond sequential order, they offer valuable inductive bias, enhancing generalization in tasks like knowledge base completion \cite{toutanova_CONV_EMNLP2015} and open-domain information extraction \cite{bowman_spinn_acl2016, angeli_openIE_IJCNLP2015}. 
Crucially for our work, these structures also improve interpretability; for example, syntax-aware attention can produce more linguistically grounded explanations \cite{li_syntax_attention_bert_ACL2021, sartran_TG_ACL2022}, and aligning model behavior with grammar improves comprehensibility of predictions \cite{mechouma_VLG_BERT_ACL2025}. While previous works typically used linguistic structures to analyze or build inherently interpretable models, we repurpose them for a post-hoc explanatory framework suitable for black-box models.

\begin{table*}[t]
\caption{\small Performance comparison on the Movie, Graph-SST, and HateXplain datasets with fixed selection rates.} %
\centering
\resizebox{1.0\textwidth}{!}{%
\begin{tabular}{l|cccc|ccc|cccc}
\toprule
\textbf{Dataset} 
&   \multicolumn{4}{c|}{\textbf{Movie}}& 
 \multicolumn{3}{c|}{\textbf{Graph-SST2}}&  \multicolumn{4}{c}{\textbf{HateXplain}}\\
\midrule
\textbf{Metric} 
&ACC&AUROC & AUPRC &   
\makecell{Annotation\\Precision}&ACC& AUROC & AUPRC& ACC&AUROC & AUPRC 
 &\makecell{Annotation\\Precision}\\
\midrule
SHAP 
& 0.491 \stdtiny{0.015} &0.505 \stdtiny{0.020}	 & 0.518 \stdtiny{0.012}  & 0.252 \stdtiny{0.003} & 0.511 \stdtiny{0.002} & 0.651 \stdtiny{0.007} & 0.631 \stdtiny{0.005} & 0.586 \stdtiny{0.002} &0.710 \stdtiny{0.004} & 0.563 \stdtiny{0.008}    &0.239 \stdtiny{0.001}\\
LIME 
& 0.495 \stdtiny{0.026}  & 0.531 \stdtiny{0.060}	 & 0.511 \stdtiny{0.032} & 0.358 \stdtiny{0.002}& 0.502 \stdtiny{0.001} & 0.611 \stdtiny{0.002}& 0.588 \stdtiny{0.003} & 0.572 \stdtiny{0.002} & 0.746 \stdtiny{0.003}	& 0.603 \stdtiny{0.004}  &0.256 \stdtiny{0.001}\\
IG 
& 0.503 \stdtiny{0.000}	 & 0.544 \stdtiny{0.000} & 0.501 \stdtiny{0.000} & 0.277 \stdtiny{0.000} & 0.500 \stdtiny{0.000} & 0.581 \stdtiny{0.000} & 0.570 \stdtiny{0.000} & 0.587 \stdtiny{0.000} & 0.761 \stdtiny{0.000} & 0.613 \stdtiny{0.000} & 0.247 \stdtiny{0.000}\\

\midrule
\rowcolor[HTML]{F5F5F5 }L2X & 0.561 \stdtiny{0.054} & 0.596 \stdtiny{0.060} & 0.590 \stdtiny{0.080} & 0.337 \stdtiny{0.021} & 0.526 \stdtiny{0.007} & 0.552 \stdtiny{0.016}	& 0.558 \stdtiny{0.016} & 0.493 \stdtiny{0.003} &0.624 \stdtiny{0.003}	& 0.479 \stdtiny{0.002}
&0.141 \stdtiny{0.002} \\
\rowcolor[HTML]{F5F5F5 }LTX & 0.589 \stdtiny{0.066} & 0.605 \stdtiny{0.078} & 0.625 \stdtiny{0.104} & 0.355 \stdtiny{0.023}
& 0.525 \stdtiny{0.007} & 0.555 \stdtiny{0.017}	& 0.563 \stdtiny{0.019} & 0.511 \stdtiny{0.003} & 0.644 \stdtiny{0.003} & 0.483\stdtiny{0.014} &0.161 \stdtiny{0.004}\\
\rowcolor[HTML]{F5F5F5 }
CXPlain & 0.512 \stdtiny{0.004} &0.571 \stdtiny{0.019} & 0.571 \stdtiny{0.010} & 0.358 \stdtiny{0.004}
& 0.524 \stdtiny{0.004} & 0.554 \stdtiny{0.016} & 0.561 \stdtiny{0.018} & 0.612 \stdtiny{0.008} &0.777 \stdtiny{0.006} & 0.645 \stdtiny{0.007} &0.234 \stdtiny{0.004}\\
\midrule
\rowcolor[HTML]{E6E6E6}MLP (HC)& 0.551 \stdtiny{0.017} & 0.590 \stdtiny{0.005} & 0.692 \stdtiny{0.004} & 0.368 \stdtiny{0.006} & 0.502 \stdtiny{0.002} & 0.512 \stdtiny{0.008} & 0.544 \stdtiny{0.006} & 0.415 \stdtiny{0.005} &0.536 \stdtiny{0.011}	& 0.369 \stdtiny{0.011} &0.181 \stdtiny{0.013}\\
\rowcolor[HTML]{E6E6E6}MLP (STE)& 0.509 \stdtiny{0.006} & 0.574 \stdtiny{0.041}	& 0.577 \stdtiny{0.044} &  0.262 \stdtiny{0.038} & 0.538 \stdtiny{0.012} & 0.577 \stdtiny{0.019} & 0.566 \stdtiny{0.011} & 0.529 \stdtiny{0.060} &0.666 \stdtiny{0.070} & 0.511 \stdtiny{0.073} &0.214 \stdtiny{0.019}\\
\rowcolor[HTML]{E6E6E6}MLP (RL)& 0.733 \stdtiny{0.015} & 0.805 \stdtiny{0.016} & 0.797 \stdtiny{0.025}  & 0.347 \stdtiny{0.011} & 0.702 \stdtiny{0.036} & 0.781 \stdtiny{0.036} & 0.770 \stdtiny{0.029} & 0.672 \stdtiny{0.003} & \textbf{0.835 \stdtiny{0.001}}& \textbf{0.713 \stdtiny{0.003}} &0.338 \stdtiny{0.002}\\

\midrule
\rowcolor[HTML]{FFF5E6}GLASS (HC)& 0.539 \stdtiny{0.009} &0.551 \stdtiny{0.008}	& 0.646 \stdtiny{0.012}	 &  0.391 \stdtiny{0.026} & 0.505 \stdtiny{0.001} & 0.513 \stdtiny{0.006}		& 0.524 \stdtiny{0.015} & 0.446 \stdtiny{0.015} &0.572 \stdtiny{0.019}	& 0.419 \stdtiny{0.022} &0.206 \stdtiny{0.007}\\
\rowcolor[HTML]{FFF5E6}GLASS (STE)& 0.605 \stdtiny{0.101} &0.639 \stdtiny{0.116} & 0.641 \stdtiny{0.119}  & \textbf{0.536 \stdtiny{0.041}}& 0.551 \stdtiny{0.015}& 0.593 \stdtiny{0.024} & 0.587 \stdtiny{0.029} & 0.550 \stdtiny{0.037} &0.677 \stdtiny{0.039} & 0.530 \stdtiny{0.035} &0.228 \stdtiny{0.011}\\
\rowcolor[HTML]{FDEBD0}\textbf{GLASS}& \textbf{0.829 \stdtiny{0.017}} & \textbf{0.954 \stdtiny{0.012}} & \textbf{0.962 \stdtiny{0.009}}  & \textbf{0.418 \stdtiny{0.005}} & \textbf{0.767 \stdtiny{0.031}} & \textbf{0.864 \stdtiny{0.022}} & \textbf{0.860 \stdtiny{0.022}} & \textbf{0.673 \stdtiny{0.002}} & 0.833 \stdtiny{0.001} & \textbf{0.713 \stdtiny{0.002}} & \textbf{0.354 \stdtiny{0.001}}\\
\midrule
\midrule
Oracle& 0.859&0.942 & 0.944  & -
 & 0.842& 0.926 & 0.928  & 0.674&0.847  & 0.740
 &-
\\
\bottomrule
\end{tabular}
}
\label{tab:main_performance_0.10} 
\end{table*}

\section{Experiment} \label{sec:experiments}
In this section, we evaluate the performance of \proposed~and cutting-edge explanation methods using several real-world datasets, complemented by comprehensive ablation studies. Additional experiments, as well as more detailed descriptions of datasets, benchmarks, and experimental settings, are provided in Appendix \ref{appendix: implementation_detail}.

\vspace{1mm}
\noindent\textbf{Benchmarks.}~~ We compare our method against seven representative explanation methods: three memoryless methods --- i.e., KernelSHAP \cite{lundberg_shap_NeurIPS2017}, LIME \cite{ribeiro_LIME_KDD2016}, and Integrated Gradients (IG) \cite{sundararajan_ig_ICML2017} --- that estimate importance without training, and four amortized method ---  i.e., L2X \cite{chen_l2x_ICML2018}, LTX\cite{barkan_LTX_ICDM2023}, CXPlain\cite{schwab_cxplain_NeurIPS2019}, INVASE\cite{yoon_invase_ICLR2018} and variants of \proposed~--- where important word subsets are inferred via a learned selector.  
To ensure a fair and rigorous comparison across a broad range of baselines, we make several improvements to the original baseline implementations. To all the amortization methods, we standardize the input features by replacing their original encoders (e.g., RNNs, CNNs, or static word embeddings like GloVe \cite{pennington_glove_EMNLP2014}) with embeddings from the same pre-trained DLM, $f$, as our method, followed by a shared MLP layer to handle variable-length input words robustly.
\footnote{This modification is done by replacing the GNN architecture in \eqref{eq:selector_notation} with an MLP layer, which is effective due to the rich contextual embeddings from $f$ that already encode complex inter-word relationships.}
It is worth highlighting that these modifications are made deliberately to ensure that all methods operate under comparable conditions, allowing each baseline to demonstrate its best possible performance within our evaluation framework.
Some model-specific details can be found below:

\begin{itemize}[leftmargin=*, topsep=0pt, partopsep=0pt, itemsep=0pt]
\item \textbf{IG}: We assume \textit{oracle access} to the gradients of the black-box DLM and use them directly to compute word-level attributions.
\item \textbf{L2X, LTX}: Since training a surrogate model that matches the predictive performance of the target black-box DLM is often infeasible, we instead treat the black-box DLM itself as the surrogate, assuming \textit{oracle access} to its gradients for selector training.

\item \textbf{Variants of \proposed:} For rigorous and compelling baselines, we introduce variants along two dimensions: (i) We replace the GNN architecture with an MLP (as described earlier), denoted as \textbf{MLP}.\footnote{This can be viewed as an extension of INVASE \cite{yoon_invase_ICLR2018} --- an instance-wise feature selection approach --- adapted for explaining black-box DLMs.} To encourage linguistic coherence, we further introduce a variant with graph Laplacian regularization, denoted as \textbf{MLP (w/LP)}, as detailed in Section~\ref{subsection:knowledge}. 
(ii) We substitute the REINFORCE algorithm with reparameterization-based methods—specifically, the Hard Concrete (\textbf{HC}) distribution~\cite{louizos_hardconcrete_ICLR2018} and the Straight-Through Estimator (\textbf{STE})~\cite{bengio_STE_2013}, under the assumption that \textit{oracle access} to the gradients of the black-box DLM is available.

\item \textbf{Oracle:} An \textit{oracle} baseline representing the black-box model's performance on full input text, serving as the target performance.
\end{itemize}

\vspace{1mm}
\noindent
\textbf{Performance Metric.}~~ We evaluate the explanation quality of each method along the two axes, i.e., discriminative/generation performance and linguistic coherence:
\begin{itemize}[leftmargin=*, topsep=0pt, partopsep=0pt, itemsep=0pt]
    \item {\textbf{Discriminative performance}:} We assess how discriminative the selected words are by assessing classification accuracy (ACC), area under the ROC curve (AUROC), and area under the precision-recall curve (AUPRC) based on the black-box predictions with selected words.
    For a fair comparison, we neutralize unselected words by either adopting the default strategy for relevant baselines (e.g., zero-embeddings for L2X) or, for all others, by replacing them with a neutral word (e.g., ``the'').
    \item {\textbf{Generation performance}:} To evaluate summary quality, we use three standard metrics: To measure lexical overlap, we use ROUGE-1 (unigrams) and ROUGE-L (longest common subsequence)~\cite{lin_rouge_ACL2004}. To capture semantic similarity, we use BERTScore \cite{Zhang_bertscore_ICLR2020}, which calculates the cosine similarity between contextual token embeddings from the generated and reference summaries, making it robust to variations in phrasing.
    
    \item {\textbf{Linguistic coherence}:} We analyze the structural properties of the subgraphs induced by the selected word subsets using external knowledge $\mathcal{K}(\mathbf{x})$. Specifically, we evaluate three metrics --- the number of subgraphs, average node degree (i.e., the average number of neighbors for each node), and edge density (i.e., the average connectivity within each subgraph) --- each reflecting the structural cohesion of the resulting explanations.
    \item {\textbf{Alignment with human annotation}:} We average the precision score (of each instance) to evaluate the proportion of words identified by the model as important that overlap with the human-annotated rationales.
\end{itemize}

\vspace{1mm}
\noindent\textbf{Implementation.}~~
For each input sentence, we construct a syntactic dependency graph using SpaCy’s dependency parser~\cite{honnibal_spacy_2020}, where each node represents a word and edges reflect grammatical relations. This provides a structured approximation of sentence syntax and enables the GNN-based selector to exploit compositional signals in natural language.
For datasets that rely solely on syntactic dependency graphs, we implement the selector $\pi_\theta$ using Graph Convolutional Networks (GCNs) \cite{kipf_gcn_ICLR2017}. 
Specifically, to ensure our selector can adapt to unique linguistic patterns of each input, we provide GCNs with instance-specific adjacency graphs, allowing the selector to flexibly adapt to varying input structures.
For the Graph-SST2 dataset, which contains denser semantic graphs, we instead adopt GraphSAGE \cite{hamilton_GraphSAGE_NeurIPS2017}, as its neighborhood aggregation mechanism is better suited for capturing rich local semantics.
Hyperparameter details can be found in Table \ref{tab:hyperparams}.

\begin{figure*}[!t]
  \centering
  \setlength{\abovecaptionskip}{0.1mm}
  \begin{subfigure}{\textwidth}
    \centering
    \includegraphics[width=1.0\linewidth]{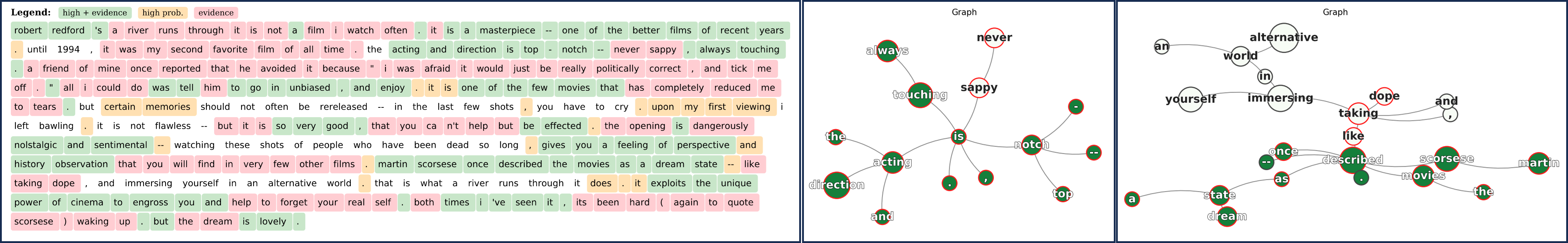}
    \caption*{\small GCN w/ syntactic dependency graph (Annotation Precision: 0.806)}
  \end{subfigure}

  \begin{subfigure}{\textwidth}
    \centering
    \includegraphics[width=1.0\linewidth]{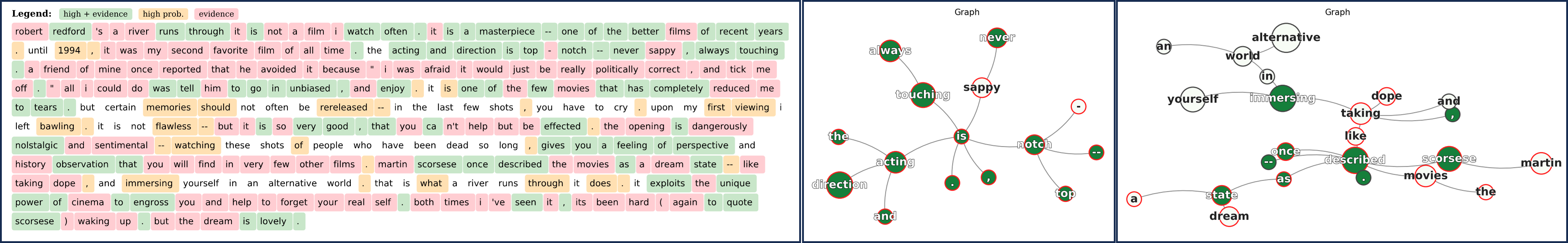}
    \caption*{\small MLP (Annotation Precision: 0.737)}
  \end{subfigure}

  \begin{subfigure}{\textwidth}
    \centering
    \includegraphics[width=1.0\linewidth]{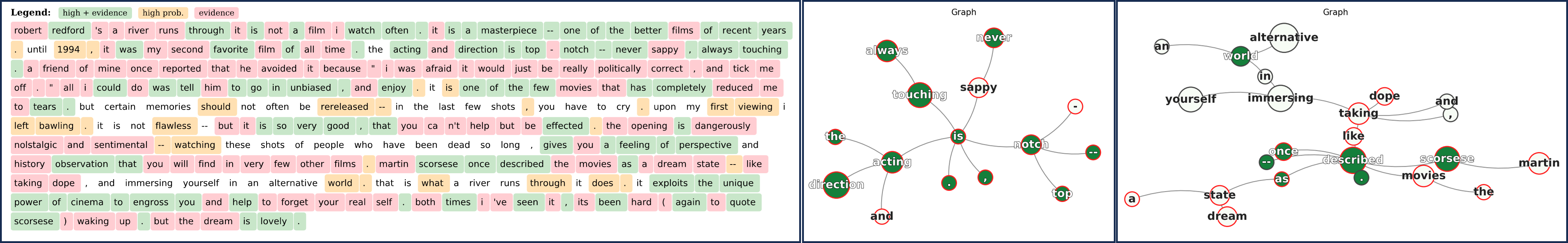}
    \caption*{\small MLP w/ graph Laplacian (Annotation Precision: 0.771)} \vspace{1mm}
  \end{subfigure}

  \caption{\small Examples of important word subsets identified by \proposed~and its variants on the Movies dataset. Based on the human rationales provided in the dataset, words highlighted in \hrationale{green} correspond to selected human rationales, those in \hmissed{red} represent unselected human rationales, and words highlighted in \hextra{orange} denote selected words that are not part of the human rationale. The accompanying subgraph illustrates the structural relationships among selected words, highlighting how well the explanation aligns with the underlying linguistic structure. We also report the sample-level precision score with respect to human rationales to quantify the degree of alignment.}
  \Description{movie_qualitative}
  \label{fig:movie_qualitative}
\vspace{-3mm}
\end{figure*}

\subsection{Sentiment Analysis}

In this subsection, we evaluate our proposed and baseline methods on two publicly available sentiment classification datasets with varying input lengths and structural complexity: 
\begin{itemize}[leftmargin=*, topsep=0pt, partopsep=0pt, itemsep=0pt]
\item \textbf{Movies} \cite{zaidan_movies_2008}: This dataset provides binary sentiment labels along with human-annotated rationales at the span level. Each review may contain multiple rationale spans that reflect different aspects or reinforce the same sentiment. These ground-truth rationales provide a reliable basis for evaluating explanation quality.
\item \textbf{Graph-SST2} \cite{yuan_graphsst2_IEEE2022}: A graph-structured version of the SST2 dataset \cite{socher_sst2_EMNLP2013}, where each sentence is augmented with a dependency parse graph constructed via the Biaffine parser~\cite{dozat_biaffine_ICLR2017}, providing the underlying semantic structure of the input sequences.
\end{itemize}

\vspace{1mm}
\noindent
\textbf{Quantitative Analysis.}~~ 
We evaluate the discriminative power of word subsets selected by our method and baseline methods under a fixed selection rate (determined by their respective importance scores). 
Table~\ref{tab:main_performance_0.10} shows that at a 10\% rate, the subsets chosen by baseline methods exhibit notably lower discriminative power than ours. 
We attribute this advantage to a fundamental design difference: while methods like KernelSHAP, IG, and LIME prioritize individually impactful words, our approach is explicitly optimized to find subsets that collectively preserve the original input's meaning and predictive signal.
This limitation of focusing on individual word impact is particularly detrimental for CXPlain since its surrogate model is trained to approximate leave-one-out importance scores.

\proposed~consistently outperforms its MLP-based variants, achieving notable gains in both the discriminative power of selected word subsets and their alignment with human-annotated rationales. This performance gain stems from the structural advantage of its GNN architecture, which effectively leverages the underlying linguistic structure of the input --- a strength that becomes especially valuable when processing longer texts.
Moreover, when comparing optimization strategies, our policy gradient method demonstrates superior performance over reparameterization tricks. We attribute this to the distinct drawbacks of the alternatives: STE introduces biased gradient estimates for its discrete selections while methods like HC, L2X, and LTX rely on soft, continuous gates for training, which leads to a train-test mismatch, as these models must make hard selections at inference, forcing them into out-of-distribution scenarios that degrade performance.

\vspace{1mm}

\noindent\textbf{Qualitative Analysis.}~~
Figure~\ref{fig:movie_qualitative} illustrates the benefit of incorporating linguistic structure by showcasing a representative example from the Movies dataset, which compares the explanation generated by our method to those from two variants: MLP and MLP w/ LP 
(those of other baselines are shown in Figure~\ref{fig:movie_qualitative_other})
. To ensure a fair comparison, all methods are configured to select an identical proportion of words (30\% of the input length):

\begin{itemize}[leftmargin=*, topsep=0pt, partopsep=0pt, itemsep=0pt]
    \item The sentence \textit{``The acting and direction is top-notch --- never sappy, always touching.''} contains a clear positive sentiment. However, MLP and MLP (w/ LP)  models independently select \textit{``never''}, \textit{``always''}, and \textit{``touching''}, yielding the misleading \textit{``never always touching''} which contradicts the true sentiment. This contradicts the positive sentiment in the full sentence and may be misinterpreted when viewed in isolation, especially since it immediately follows the strongly positive phrase \textit{``acting and direction is top-notch''}. 
    Conversely, our GCN-based selector identifies coherent, structure- and meaning-preserving phrases—\textit{acting and direction is top-notch''} and \textit{always touching''}—that better align with the original sentiment.
    \item The sentence \textit{``Martin Scorsese once described the movies as a dream state --- like taking dope, and immersing yourself in an alternative world.''} illustrates a failure case for the MLP and MLP (w/ LP) models, which produce fragmented and semantically sparse word selections that offer little interpretive value. This is due to their inability to model local structure or capture long-range dependencies.
    Contrarily, our GCN-based selector identifies a cohesive span of structurally connected words that align with the dependency graph, producing a fluent and interpretable phrase. The selected subspan effectively captures the themes of escapism and immersion, and aligns well with the sentence's sentiment label. This demonstrates the advantage of incorporating linguistic structure into the selection process.
\end{itemize}

\subsection{Explainable Hate Speech Detection}

To demonstrate applicability in the nuanced and often ambiguous context of social media language \cite{mathew_hatexplain_AAAI2021, davidson_hatespeech_AAAI2017}, we evaluate our method and the baseline methods in the task of hate speech detection.
\begin{itemize}[leftmargin=*, topsep=0pt, partopsep=0pt, itemsep=0pt]
    \item \textbf{HateXplain}~\cite{mathew_hatexplain_AAAI2021}: 
    This dataset focuses on hate speech and includes multi-faceted annotations for each post, including toxicity level (hate, offensive, or normal), the targeted group, and crucially, human-annotated rationales (i.e., text spans that justify the assigned label). The presence of these ground-truth rationales makes it an ideal dataset for evaluating explanation fidelity. %
\end{itemize}

\vspace{1mm}

\noindent 
\textbf{Quantitative Results.}~~
While \proposed~significantly outperforms conventional methods on this dataset in discriminating toxicity levels, its performance on this metric is comparable to its MLP-based variant --- a result that differs from our findings on the Movies and Graph-SST2 datasets. We conjecture this is due to the dataset's informal language and community-specific slang, characteristics known to reduce the effectiveness of the formal structural biases our GNN relies on \cite{eisenstein_badlanguage_NAACL2013, baldwin_socialtext_IJCNLP2013}. Nevertheless, even with comparable discriminative power, \proposed~produces superior explanations. Metrics measuring precision against human-annotated rationales and linguistic coherence both confirm that our GNN-based method captures structural information more effectively than the MLP variant. 
See Figure \ref{fig:Word_subsets_explanation_hatexplain} for qualitative examples.

\subsection{Abstractive Text Summarization} \label{main: summarization}

To demonstrate the broader applicability of our method beyond classification, we evaluate it on a challenging generative benchmark known for its high level of abstraction. 
Our approach offers key flexibility by learning from appropriately defined reward signals without task-specific constraints.

\begin{itemize}[leftmargin=*, topsep=0pt, partopsep=0pt, itemsep=0pt]
    \item \textbf{XSum}~\cite{narayan_xsum_emnlp2018}: This dataset is a large-scale collection of BBC articles, each with a single-sentence summary that abstractly describes the article's main topic. Since the ground-truth summaries have minimal lexical overlap with the original text, this makes it an ideal dataset for evaluating explanation methods for true language understanding and novel text generation.
\end{itemize}

\noindent
\textbf{Quantitative Results.}~~
Even in this challenging abstractive summarization task, where low lexical overlap makes selection-based methods difficult, \proposed~demonstrates robust performance. Despite the inherent constraint of using only a partial input (i.e., selection rate of 50\%), a notable finding emerges: while lexical overlap metrics like ROUGE-L expectedly decrease, the semantic-based BERTScore remains high. This indicates that our selector is adept at identifying and preserving the core semantic meaning of a text even with a limited word subset, which is crucial for effective abstractive summarization. 
See Figure \ref{fig:Word_subsets_explanation_xsum} for qualitative examples.

\subsection{Ablation Studies}

\begin{table}[t]
\centering
\caption{\small Generation performance for the XSum dataset.} %

\label{tab:xsum_results}
\resizebox{0.65\columnwidth}{!}{
    \begin{tabular}{l|c|c|c}
    \toprule
    \textbf{Dataset} & \multicolumn{3}{c}{\textbf{XSum}} \\
    \midrule
    \textbf{Metric} & \textbf{Rouge-1}& \textbf{Rouge-L} & \textbf{Bert Score} \\
    \midrule
    \rowcolor[HTML]{E6E6E6}MLP (RL) & $0.245$ & $0.196$ & $0.849$ \\
    \rowcolor[HTML]{FDEBD0}\textbf{GLASS}& $0.277$ & $0.219$  & $0.882$ \\
    Oracle & $0.391$ & $ 0.320$ & $0.908$ \\
    \bottomrule
    \end{tabular}
}
\end{table}

\noindent \textbf{Linguistic Structures.}~~
To further demonstrate the effectiveness of our method in leveraging GNN architectures with input-specific external knowledge $\mathcal{K}(\mathbf{x})$, we evaluate how well the selected word subsets align with the underlying linguistic structure. We compare our GCN-based selector with two variants of our method: MLP and MLP (w/ LP) (see Appendix \ref{appendix: Laplacian} for details). 
This comparison relies on coherence metrics that gauge faithfulness to the syntactic prior by assessing if the selected words: (i) form cohesive linguistic units (i.e., phrases or clauses), as measured by subgraph count and edge density, and (ii) include pivotal grammatical elements, as indicated by average node degree.

\begin{figure}[t] 
\centering
\includegraphics[ width=\linewidth]{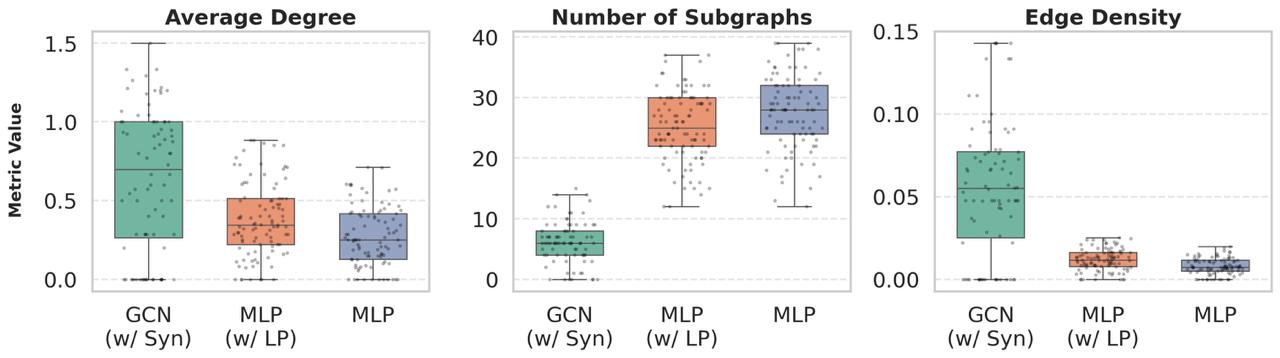} %
\caption{\small Comparison of linguistic coherence with respect to (left) average node degree, (middle) the number of subgraphs, and (right) edge density in the Movies dataset.} 
\Description{structure_coherence}
\label{fig:structure_coherence}
\vspace{-3mm}
\end{figure} 

\textbf{\textit{Why is the structural coherence so important?}} The pursuit of structural coherence in our explanations is not a byproduct but a central design choice aimed at improving comprehensibility~\cite{embick_activationArea_PNAS2000, MULLER_LinguisticfMRI_Neuropsychologia2003}. This approach is motivated by the cognitive principle that humans process cohesive informational units more effectively than scattered individual words~\cite{Matthew_chunk_PNAS2017}. Our coherence metrics are designed as direct proxies for this goal. Specifically, a low subgraph count and high edge density indicate that selected words form cohesive "clusters" rather than isolated "islands". Additionally, a high average node degree anchors explanations around the sentence's grammatical "hubs" (e.g., the main verb), thus preserving its core structure. 
As presented in Figure~\ref{fig:structure_coherence}, our GCN-based approach consistently excels in these metrics when selecting approximately 10\% of the words, confirming its ability to produce structurally coherent explanations.

\noindent \textbf{External Knowledge.}~~ 
We also examine how different types of linguistic graphs, $\mathcal{K}(\mathbf{x})$, used as external knowledge, influence performance. On the Graph-SST2 dataset, we compare our selector’s performance when guided by the dataset-provided semantic graphs versus conventional syntactic dependency graphs. As shown in Table~\ref{tab:sst2_graph_type_10}, both graph types substantially outperform MLP baselines, highlighting the effectiveness of structured linguistic guidance. Notably, the semantic graph yields a slight performance gain over the syntactic graph, suggesting that richer linguistic priors can further enhance the quality of the selected explanations.

\begin{table}[!t] 
\centering
\caption{\small Performance comparison with different graph types and selector architectures on the Graph-SST2 dataset.} %

\label{tab:sst2_graph_type_10}
\resizebox{\linewidth}{!}{%
\begin{tabular}{l|ccc}\toprule
 \textbf{Graph Type}& ACC & AUROC & AUPRC \\
\midrule
 MLP
& 0.702 \stdtiny{0.036} & 0.781 \stdtiny{0.036} & 0.770 \stdtiny{0.029} \\
 \rowcolor[HTML]{F2F2F2}MLP (w/ Syn LP)& 0.724 \stdtiny{0.038} & 0.793 \stdtiny{0.033} & 0.760 \stdtiny{0.029} \\
 \rowcolor[HTML]{F2F2F2}MLP (w/ Sem LP)& 0.724 \stdtiny{0.036} & 0.791 \stdtiny{0.019} & 0.743 \stdtiny{0.010} \\
 \rowcolor[HTML]{F2F2F2}MLP (w/ Syn \& Sem LP)& 0.713 \stdtiny{0.027} & 0.786 \stdtiny{0.020} & 0.761 \stdtiny{0.019} \\
 \rowcolor[HTML]{FDEBD0}GNN (w/ Syn)& 0.767 \stdtiny{0.031} & 0.864 \stdtiny{0.022} & 0.860 \stdtiny{0.022} \\
 \rowcolor[HTML]{FDEBD0}GNN (w/ Sem)& \textbf{0.788 \stdtiny{0.021}}& \textbf{0.884 \stdtiny{0.015}}& \textbf{0.880 \stdtiny{0.013}}\\
 \rowcolor[HTML]{FDEBD0}GNN (w/ Syn \& Sem)& 0.784 \stdtiny{0.025} & 0.881 \stdtiny{0.020} & 0.879 \stdtiny{0.015} \\
\midrule
Oracle& 0.842& 0.926& 0.928\\
\bottomrule
\end{tabular}%
}
\vspace{-4mm}
\end{table}

\vspace{1mm}
\noindent \textbf{Model Configurations.}~~
In this ablation study, we assess the model dependency of our framework by training selectors with various embedding models ($f$) for a diverse set of target black-box DLMs ($g$). As presented in Table \ref{tab:encoder_selector_comparison_010}, our approach demonstrates high flexibility, successfully producing robust selectors for a wide range of target architectures, including both encoder-based (e.g., DeBERTa) and decoder-based (e.g., GPT-2~\cite{radford_GPT2_OpenAI2019}) models, using various types of embedding models. 
This confirms that the observed performance gain stems not from architectural similarity between $f$ and $g$, but from our proposed framework itself. 
Further results for different embedding model sizes are available in Table \ref{tab:deberta_size_comparison}.
 
\subsection{Additional Experiments}
\textbf{Generalization.}~~
To evaluate the selector's generalization ability and ensure it avoids overfitting to source dataset characteristics, we train the selector exclusively on the Movies dataset and apply it to the IMDb dataset \cite{maas_imdb_acl2011} --- which addresses the same movie review sentiment task but is 25 times larger --- without any fine-tuning or adaptation.
As shown in Table~\ref{tab:cross_dataset_metrics}, the selector maintains a high performance. This result indicates that our selector learns robust, task-relevant features, functioning effectively as a task-level explainer that transfers across datasets with little performance loss.

\vspace{1mm}
\noindent
\textbf{API-Based Model Compatibility.}~~ 
To validate real-world applicability, we evaluate GLASS on keyword extraction using the Inspec dataset \cite{hulth_inspec_EMNLP2003} with a Hugging Face API-based model \cite{kulkarni_inspecBERT_2022}, under strict input-output access only. As shown in Table~\ref{tab:inspec}, GLASS achieves an F1 score of 0.441 ± 0.004 using 30\% of the input text, which is 90.2\% of oracle performance (F1 score: 0.489), demonstrating full compatibility with production-level API-based models, as detailed in Appendix~\ref{appendix:api_expriments}.

\vspace{1mm}
\noindent
\textbf{Scalability to Larger DLMs.}~~
To evaluate GLASS on larger models, we conduct experiments using Llama 3.2 (3B) as the target black-box model ($g$) while using the much smaller BERT-base (110M) as the embedding model ($f$). GLASS achieves competitive performance across all metrics on the Movies dataset, even outperforming the full-text black-box model as shown in Table~\ref{tab:lamma3}. Notably, our method achieves these results using only 10\% of the original input words, demonstrating its robust performance for explaining larger DLMs.

\begin{table}[t]
\centering
\caption{\small Performance comparison with different encoder-selector combinations on the Graph-SST2 dataset.
}%

\label{tab:encoder_selector_comparison_010}
\resizebox{\linewidth}{!}{%
\begin{tabular}{cc|ccc}\toprule
\textbf{$f$}& \textbf{$g$}& ACC & AUROC & AUPRC \\
\midrule
BERT & & 0.767 \stdtiny{0.031} & 0.864 \stdtiny{0.022}	 & 0.860 \stdtiny{0.022} \\
DeBERTa-v3 & DeBERTa-v3
& 0.840 \stdtiny{0.006} & 0.930 \stdtiny{0.006} & 0.925 \stdtiny{0.009} \\
RoBERTa & & 0.823 \stdtiny{0.003} & 0.920 \stdtiny{0.004}	 & 0.917 \stdtiny{0.005} \\
\midrule
BERT & & 0.827 \stdtiny{0.006} & 0.911 \stdtiny{0.002} & 0.913 \stdtiny{0.003}\\
DeBERTa-v3 & GPT-2
& 0.887 \stdtiny{0.003} & 0.950 \stdtiny{0.003} & 0.950 \stdtiny{0.004}\\
RoBERTa & & 0.869 \stdtiny{0.007}	& 0.940 \stdtiny{0.004}	& 0.941 \stdtiny{0.005}  \\
\bottomrule
\midrule
\multirow{2}{*}{Oracle}& DeBERTa-v3 & 0.842& 0.926& 0.928\\
& GPT-2      & 0.819& 0.907& 0.913\\

\bottomrule
\end{tabular}
}
\rule{0pt}{0pt}
\vspace{-2mm}
\end{table}

\begin{table}[t]
\centering
\caption{\small Evaluating Transfer Performance from \textbf{Movies} to \textbf{IMDb}.} %

\label{tab:cross_dataset_metrics}
\resizebox{\columnwidth}{!}{%
\begin{tabular}{@{}l|cc|cc|cc@{}}
\toprule
\textbf{$\pi_\theta \to g$} & \multicolumn{2}{c|}{\textbf{Movies $\to$ Movies}} & \multicolumn{2}{c|}{\textbf{IMDb $\to$ IMDb}} & \multicolumn{2}{c}{\textbf{Movies $\to$ IMDb}} \\
\midrule
\textbf{Metric} & \textbf{AUROC} & \textbf{AUPRC} & \textbf{AUROC} & \textbf{AUPRC} & \textbf{AUROC} & \textbf{AUPRC} \\
\midrule
\rowcolor[HTML]{FDEBD0}& 0.952 & 0.961 & 0.940 & 0.936 & 0.937 & 0.949 \\
\rowcolor[HTML]{FDEBD0}\multirow{-2}{*}{\textbf{GLASS}}& \stdtiny{0.013} & \stdtiny{0.010} & \stdtiny{0.006} & \stdtiny{0.006} & \stdtiny{0.004} & \stdtiny{0.003} \\
Oracle& 0.942 & 0.944 & 0.987 & 0.996 & 0.987 & 0.996 \\
\bottomrule
\end{tabular}
}
\vspace{-2mm}
\end{table}

\begin{table}[t]
\centering
\caption{\small Keyword extraction performance on Inspec dataset using API-based model.}
\label{tab:inspec}
\begin{tabular}{lcc}
\toprule
Method  & F1 Score \\
\midrule
\rowcolor[HTML]{FDEBD0} GLASS & 0.441 $\pm$ 0.004 \\
Oracle (Full Text) & 0.489 \\
\bottomrule
\end{tabular}
\vspace{-2mm}
\end{table}

\begin{table}[!t]
    \centering
    \caption{\small GLASS performance on the Movies dataset using Llama 3.2 as the target black-box model.} %

    \label{tab:lamma3}
    \resizebox{0.9\columnwidth}{!}{
    \begin{tabular}{l|ccc}
        \toprule
        Model & ACC & AUROC & AUPRC \\
        \midrule
        \rowcolor[HTML]{FDEBD0}\textbf{GLASS}& 0.774 \stdtiny{0.017} & 0.876 \stdtiny{0.015} & 0.877 \stdtiny{0.016} \\
        Llama3.2 (Oracle) & 0.759 & 0.850 & 0.862 \\
        \bottomrule
    \end{tabular}
    }
\vspace{-2mm}
\end{table}

\vspace{1mm}
\noindent
\textbf{Computational Efficiency.}~~
As shown in Table~\ref{tab:efficiency}, GLASS generates explanations via one-shot inference, taking only 0.08s per sample on average. To ensure a fair comparison with memoryless methods, all experiments are conducted with a batch size of 1. For 199 samples using DeBERTa-v3. Importantly, while the query cost of LIME and KernelSHAP scales prohibitively with model size ($g$), GLASS maintains high efficiency. 
As shown in Table~\ref{tab:efficiency}, GLASS generates explanations via one-shot inference, taking only 0.08s per sample---8--77$\times$ faster than memoryless methods---while consuming less GPU memory (6.2\,GB vs.\ 7.3--28.2\,GB). All experiments use batch size 1 for fair comparison.

\begin{table}[t]
    \centering
    \caption{\small Efficiency evaluation in terms of inference time per sample and peak GPU memory usage.}
    \label{tab:efficiency}
    \resizebox{\columnwidth}{!}{
    \begin{tabular}{ll|cc}
        \toprule
        \textbf{Type} & \textbf{Method} & \textbf{Time/Sample (s)} 
        & \textbf{Peak GPU Mem (GB)} \\
        \midrule
        \multirow{3}{*}{\textbf{Memoryless}} 
        & KernelSHAP & 2.00 & 7.35 \\
        & LIME        & 6.13 & 7.39 \\
        & IG          & 0.67 & 28.25 \\
        \midrule
        \multirow{2}{*}{\textbf{Amortized}}
        & \cellcolor[HTML]{FFF5E6}MLP (RL) 
          & \cellcolor[HTML]{FFF5E6}0.08 
          & \cellcolor[HTML]{FFF5E6}6.23 \\
        & \cellcolor[HTML]{FDEBD0}GLASS         
          & \cellcolor[HTML]{FDEBD0}0.08 
          & \cellcolor[HTML]{FDEBD0}6.24 \\
        \bottomrule
    \end{tabular}
    }
\end{table}

\vspace{1mm}
\noindent
\textbf{Human Evaluation.}~~
Since evaluating explanations can be inherently subjective, we complement our automatic metrics with a human evaluation involving 26 AI researchers. Participants rank explanations from \proposed, MLP~(RL), and IG along two dimensions: (1)~\textit{Rationale Quality} --- given the model's prediction, which selected words are most essential? --- and (2)~\textit{Phrase Coherence} --- which selections form more natural, coherent phrases rather than scattered words? Method identities are anonymized and option ordering is randomized to reduce potential bias. As shown in Table~\ref{tab:human_eval}, \proposed~receives the most Rank-1 preferences on both dimensions, with all differences statistically significant ($p < 0.0001$, Wilcoxon signed-rank test with Bonferroni correction), providing strong evidence that \proposed~produces explanations that are both more faithful and linguistically coherent from a human perspective.

\begin{table}[t]
\caption{\small Human preference distribution (Rank 1 to 3) evaluating Rationale Quality and Phrase Coherence across different explanation models.}
\label{tab:human_eval}
\resizebox{\columnwidth}{!}{
\begin{tabular}{l|ccc|ccc}
\toprule
\multirow{2}{*}{\textbf{Method}} 
  & \multicolumn{3}{c|}{\textbf{Rationale Quality}} 
  & \multicolumn{3}{c}{\textbf{Phrase Coherence}} \\
  &  Rank-1 & Rank-2 & Rank-3 
  & Rank-1 & Rank-2 & Rank-3 \\
\midrule
\rowcolor[HTML]{FDEBD0}\textbf{GLASS} 
  & \textbf{275} & 147 & 98 
  & \textbf{395} & 86  & 39  \\
MLP (RL)       
  & 135 & \textbf{251} & 134 
  & 83  & \textbf{345} & 92  \\
IG             
  & 110 & 122 & \textbf{288} 
  & 42  & 89  & \textbf{389} \\
\bottomrule
\end{tabular}
}
\end{table}

\vspace{1mm}
\noindent\textbf{Comparison with LLM-based Methods.}~~ We compare GLASS against recent LLM-based prompting methods~\cite{kroeger_counterfactualLLM_2025} via three strategies: \textit{Direct Prompting}~(DP), \textit{Counterfactual-Parallel}~(CFP), and \textit{Counterfactual\allowbreak-\allowbreak Sequential}~(CFS). While self-explanation methods benefit from LLMs' broad knowledge, they rely on the model to verbalize its own reasoning, which may not faithfully reflect its actual decision process~\cite{parcalabescu_selffaithful_2024}. As shown in Table~\ref{tab:llm_comparison}, \proposed~outperforms all LLM-based methods across all metrics despite using a substantially smaller embedding model (DeBERTa-v3 vs.\ Llama3~80B or GPT-4o).

\begin{table}[t]
\centering
\caption{\small Comparison with LLM-based prompting methods on the 
Movies dataset.} %
\label{tab:llm_comparison}
\resizebox{\columnwidth}{!}{
\begin{tabular}{ll|cccc}
\toprule
\textbf{Model} & \textbf{Method} & \textbf{ACC} & \textbf{AUROC} 
& \textbf{AUPRC} & \textbf{\makecell{Annotation\\Precision}} \\
\midrule
\multirow{3}{*}{Llama3 80B} 
  & DP  & 0.814 & 0.907 & 0.841 & 0.244 \\
  & CFP & 0.799 & 0.900 & 0.845 & 0.216 \\
  & CFS & 0.814 & 0.889 & 0.795 & 0.243 \\
\midrule
\multirow{3}{*}{GPT-4o}     
  & DP  & 0.749 & 0.916 & 0.896 & 0.245 \\
  & CFP & 0.533 & 0.809 & 0.705 & 0.224 \\
  & CFS & 0.764 & 0.913 & 0.864 & 0.236 \\
\midrule
\rowcolor[HTML]{FDEBD0}DeBERTa-v3 
  & \textbf{GLASS} 
  & \textbf{0.829} & \textbf{0.954} & \textbf{0.962} & \textbf{0.418} \\
\bottomrule
\end{tabular}
}
\end{table}

\section{Conclusion} \label{sec:conclusion}
We propose a novel method for explaining \textit{strict} black-box DLMs that combines amortized optimization with linguistic structural knowledge. Our innovations --- amortized optimization with policy gradient training for discrete selection without internal model access and the direct incorporation of linguistic structural knowledge --- enable faithful and coherent word subset explanations. Empirical results confirm that our approach significantly surpasses existing methods, even those with privileged gradient access, in identifying discriminative and linguistically sound rationales. This work contributes a practical and effective solution for increasing the transparency and trustworthiness of DLMs in critical applications.

\vspace{1mm}
\noindent
\textbf{Limitations.}~~
While our method demonstrates strong performance and practical utility in strict black-box settings, some limitations remain. First, our framework assumes that external knowledge is provided in the form of structured linguistic graphs; incorporating other modalities (e.g., tabular metadata) would require additional architectural modifications to effectively represent and utilize such information. 
Second, if users wish to incorporate new forms of structural knowledge into the explanation process, our method requires retraining the selector to integrate this knowledge, posing a practical constraint in dynamic or user-driven applications. 
Third, our approach relies on the quality of the provided linguistic graph and may vary in domains with noisy text or specialized language where standard parsers generate less informative structures.

\begin{acks}
We thank the reviewers for their comments and suggestions. This work is supported by the National Research Foundation of Korea (NRF) grant funded by the Korea government (MSIT) (No. RS-2024-00358602) and the Institute of Information \& Communications Technology Planning \& Evaluation (IITP) grant funded by the Korea government (MSIT), Artificial Intelligence Graduate School Program (No. RS-2019-II190079, Korea University), the Artificial Intelligence Star Fellowship Support Program to nurture the best talents (No. RS-2025-02304828),  and the AI Research Hub Project (No. RS-2024-00457882). 
\end{acks}

\bibliographystyle{ACM-Reference-Format}
\bibliography{reference}

\appendix

\renewcommand{\thetable}{S.\arabic{table}}
\renewcommand{\thefigure}{S.\arabic{figure}}
\renewcommand{\theequation}{\alph{section}.\arabic{equation}}
\setcounter{table}{0}
\setcounter{figure}{0}
\setcounter{equation}{0}

\section{Details about Formal Derivations}

\subsection{Extension to Regression Tasks} \label{appendix: regression}
To extend our framework to regression tasks, we modify the loss $\ell_y$ to reflect the change in the prediction error based on the black-box DLM given an entire sentence and that given the selected word subsets as follows: 
\begin{equation}
    \ell_y(g(\mathbf{x}), g(\tilde{\mathbf{x}})) = -\left\|y - g(\mathbf{x})\right\|_2 + \left\|y - g(\tilde{\mathbf{x}})\right\|_2
\end{equation}
where $y \in \mathbb{R}$ is a continuous target label and $g: \mathcal{X} \rightarrow \mathbb{R}$ is a regressor based on the black-box DLM. 

 \subsection{Derivation of (\ref{eq:Reinforce_loss})}\label{Reinforce_loss_derivation}
 Applying the policy gradient method, we can compute the gradient of our objective $L(\theta)$ with respect to the selector parameters $\theta$ as follows, where we define the term $R(\mathbf{x}, \mathbf{m})$ as:
\begin{equation} \label{eq:R_term}
    R(\mathbf{x}, \mathbf{m}) = \ell_y\big(g(\mathbf{x}),g(\tilde{\mathbf{x}}) \big) + \lambda||\mathbf{m}||_0.
\end{equation}
The derivation proceeds as:
\begin{align} \label{eq:Reinforce_loss_derivation}
    \nabla_\theta \mathcal{L}(\theta)
    &= \mathbb{E}_{\mathbf{x}, y\sim p_{XY}}  \left[\sum_{\mathbf{m} \in \{0,1\}^T} \nabla_\theta p_{\theta}(\mathbf{m}) \cdot R(\mathbf{x}, \mathbf{m}) \right] \nonumber \\
    &= \mathbb{E}_{\mathbf{x}, y\sim p_{XY}}  \left[\sum_{\mathbf{m} \in \{0,1\}^T} p_{\theta}(\mathbf{m}) \frac{\nabla_\theta p_{\theta}(\mathbf{m})}{p_{\theta}(\mathbf{m})} \cdot R(\mathbf{x}, \mathbf{m}) \right] \nonumber \\
    &= \mathbb{E}_{\mathbf{x}, y\sim p_{XY}} \Big[ \mathbb{E}_{\mathbf{m} \sim \texttt{Bern}(\pi_{\theta}(\mathbf{x}))} \Big[ \nabla_\theta \log p_{\theta}(\mathbf{m}) \cdot R(\mathbf{x}, \mathbf{m}) \Big] \Big]
\end{align}

\subsection{Graph Laplacian} \label{appendix: Laplacian}

In the regularization-based approach, the linguistic structure of a knowledge graph can be directly utilized as a regularizer on the selector's output. A common method involves applying graph Laplacian regularization, which encourages the assignment of similar selection probabilities to structurally connected words. Suppose we are given an adjacency matrix ($A \in \mathbb{R}_{\geq 0}^{T\times T}$) based on linguistic structure, we can compute the graph Laplacian as follows:
\begin{equation} \label{eq:laplacian_regularization}
    \mathcal{L}_{\text{reg}}(\mathbf{x}, A)  = \sum_{t,t'} A_{tt'} \left(\pi_\theta(\textbf{x})_t - \pi_\theta(\textbf{x})_{t'}\right)^2 = 2 \cdot \pi_\theta(\mathbf{x})^\top L \pi_\theta(\mathbf{x}),
\end{equation}
where $L = D - A$ represents the unnormalized graph Laplacian constructed based on the adjacency matrix $A$ and the degree matrix $D_{tt} = \sum_{t'}A_{tt'}$. 

While Laplacian regularization effectively smooths selector outputs according to the linguistic structure of a given input sequence, it relies on a strong assumption that edge connectivity directly translates to functional similarity.
However, as noted in prior works \cite{kipf_gcn_ICLR2017}, graph edges often encode diverse relationships --- such as contrastive, hierarchical, or task-specific dependencies --- that do not inherently guarantee such similarity. 
Applying Laplacian regularization to enforce similarity between syntactically linked but semantically contrasting factors would therefore obscure their distinct impacts, diminishing the clarity and interpretability of the resulting explanations.

\section{Implementation Details} \label{appendix: implementation_detail}

We perform a grid search to tune the hyperparameters for both \textbf{GLASS} and all baselines. Due to the instability inherent in policy gradient methods, careful tuning is required for each dataset individually. We follow the standard train/validation/test splits as provided by each dataset.

For each task, we fine-tune a pretrained DLM with an added classification head to serve as the black-box predictor. After fine-tuning, this model is treated as a strictly black-box model (denoted as $g$), with no access to internal components such as gradients or embeddings.

To initialize the selector network, we use contextual embeddings extracted from a pretrained DLM (denoted as $f$), which is \emph{not} fine-tuned and remains fixed throughout. 
This word embedding model $f$ is deliberately kept separate from the target black-box model and is used solely as a fixed (frozen) feature extractor for the selector during its training. The choice of $f$ is guided by its capacity to represent rich inter-word relationships adequately, thereby enabling effective and efficient feature extraction without introducing new trainable parameters into our system. We use the same architecture as that of the black-box DLM $g$ as a baseline, but also experiment with various DLMs as $f$ to assess architectural influences. By default, these embeddings are extracted from the 10th layer of a 12-layer transformer, though we also evaluate embeddings from the initial, 5th, and final layers for comparison. Given that DLMs typically operate on subword tokens, we aggregate these into word-level representations via averaging. For enhanced training efficiency, all embeddings are precomputed and cached.

Training \textbf{GLASS} takes approximately 3 hours on the Movies dataset~\cite{zaidan_movies_2008} (40 epochs), 8 hours on Graph-SST2~\cite{yuan_graphsst2_IEEE2022} (40 epochs), 3 hours on IMDb~\cite{maas_imdb_acl2011} (5 epochs), and 2.5 hours on HateXplain~\cite{mathew_hatexplain_AAAI2021} (30 epochs). All experiments are conducted on a single GPU machine\footnote{CPU: Intel Xeon Gold 6526Y (16 cores, 32 threads); GPU: NVIDIA A6000 (48GB VRAM).}.

\subsection{GLASS} \label{appendix: glass}
Training \textbf{GLASS} requires careful hyperparameter tuning due to the instability of policy gradient methods.  Despite this, we consistently identified effective configurations across all datasets and encoder DLMs. These selected training coefficients for each setting are summarized in Table~\ref{tab:hyperparams}.

To further reduce variance during training, we move the \(\ell_0\) regularization term in our objective outside the inner expectation by replacing it with the \(\ell_1\) norm of the selection probabilities. Hence, the gradient of our objective $\mathcal{L}(\theta)$ becomes: 
\begin{align} \label{eq:selector_loss}
    \nabla_{\theta} \mathcal{L}(\theta)
    &=
    \begin{aligned}[t] %
        &\mathbb{E}_{ \mathbf{x},y \sim p_{XY}}\Big[\mathbb{E}_{\mathbf{m} \sim \texttt{Bern}(\pi_{\theta}(\mathbf{x}))}\big[ \nabla_\theta \log p_{\theta}(\mathbf{m}) \cdot \big( \ell_y\big(g(\mathbf{x}),g(\tilde{\mathbf{x}}) \big) \\
        &\qquad + \lambda||\mathbf{m}||_0 \big) \big]\Big] \nonumber
    \end{aligned} \\
    &= 
    \begin{aligned}[t] %
        &\mathbb{E}_{\mathbf{x},y \sim p_{XY}}\big[\mathbb{E}_{\mathbf{m} \sim \texttt{Bern}(\pi_{\theta}(\mathbf{x}))}[ \nabla_\theta \log p_{\theta}(\mathbf{m})  \cdot  \ell_y\big(g(\mathbf{x}),g(\tilde{\mathbf{x}}) \big)] \\
        &\qquad + \nabla_{\theta} \lambda|| \pi_\theta(\mathbf{x})||_1\big]
    \end{aligned}
\end{align}

\begin{table*}[t!] 
\centering
\caption{Hyperparameters of \textbf{GLASS}} 
\label{tab:hyperparams}
\begin{tabular}{@{}ll@{}}
\toprule
\textbf{Block} & \textbf{Set of Hyperparameters} \\
\midrule
Initialization & Xavier~\cite{glorot_Xavier_AISTATS2010} \\
Optimization & Adam~\cite{kingma_adam_ICLR2014} \\
Mini-batch size & 128 (Graph-SST2, HateXplain), 32 (Movies, IMDb, XSum, Inspec) \\
Non-linearity & ReLU \\
\midrule
\multicolumn{2}{@{}l}{\textbf{Selector Training}} \\
Learning rate ($\eta$) & \{0.0001, 0.00025, 0.0005, 0.001\} \\
No. of hidden units & \{10, 30, 50, 100\} \\
No. of GNN layers & 2 \\
No. of encoder layers & 
\begin{tabular}[t]{@{}l@{}}
10 (BERT~\cite{devlin_Bert_NAACL2019}, DeBERTa~\cite{he_deberta_ICLR2021}, RoBERTa~\cite{liu_roberta_arXiv2019}, BART~\cite{lewis_bart_ACL2020}), 4 (DistilBERT~\cite{kulkarni_inspecBERT_2022})
\end{tabular}\\
 No. of sampling& 1 (XSUM), 8 (Other)\\
* Masking strategy&Replace with ``the'' or remove unselected words\\
Coeff. $\lambda$ &  \{0.00625 (Movies), 0.002 (IMDb), 0.01 (Graph-SST2), 0.02 (HateXplain), 0.1 (XSum, Inspec)\} \\
\bottomrule
\end{tabular}
\end{table*}

\begin{table*}[t]
\centering
\caption{Summary of datasets used in our experiments, including splits and task type.}
\label{tab:datasets}
\resizebox{0.8\textwidth}{!}{%
\begin{tabular}{@{}lcccc@{}}
\toprule
\textbf{Dataset} & \textbf{Train} & \textbf{Valid} & \textbf{Test} & \textbf{Task Type} \\
\midrule
Movies\textsuperscript{*}~\cite{zaidan_movies_2008} & 1,600 & 200 & 199 & Document-level sentiment classification\\
IMDb~\cite{maas_imdb_acl2011}& 22,500 & 2,500 & 25,000 & Document-level sentiment classification \\
HateXplain\textsuperscript{*}~\cite{mathew_hatexplain_AAAI2021} & 15,383 & 1,922 & 1,924 & Multi-aspect hate speech classification (hate/offensive/normal) \\
Graph-SST2\textsuperscript{\dag}~\cite{yuan_graphsst2_IEEE2022} & 67,348 & 872 & 1,821 & Sentence-level sentiment classification \\
XSum~\cite{yuan_graphsst2_IEEE2022} & 204,017 & 11,327 & 11,333 & Abstractive Text Summarization\\
Inspec~\cite{hulth_inspec_EMNLP2003} & 1,000 & 500 & 500 & Keyword extraction\\

\bottomrule
\end{tabular}%
}

\textsuperscript{*}Includes human-annotated rationales. ~~~~
\textsuperscript{\dag}Includes semantic graph inputs.
\end{table*}

\subsection{Extension to multiple Knowledge integration} \label{appendix: extension_knowledge}
We provide an extension of our method that leverages multiple forms of external knowledge. To accommodate this, we design our selector to flexibly incorporate multiple structural graphs without requiring architectural modification. Specifically, we adopt a modular GNN architecture that processes each structural input independently and aggregates the resulting outputs. Let \(\{\mathcal{K}^{(k)}(\mathbf{x})\}_{k=1}^K\) denote a collection of graph-structured knowledge sources, where each \(\mathcal{K}^{(k)}(\mathbf{x})\) encodes a distinct relational view over the input sequence (e.g., syntactic or semantic dependencies). Given shared node features \(f(\mathbf{x}) \in \mathbb{R}^{T \times d}\), we compute:
\begin{align}
    \mathbf{h}^{(k)} &= \texttt{GNN}^{(k)}(f(\mathbf{x}), \mathcal{K}^{(k)}(\mathbf{x})), \quad\text{for } k = 1,\dots,K \\
\mathbf{h} &= \texttt{MLP}(\mathbf{h}^{(1)}, \dots, \mathbf{h}^{(K)}), \quad \mathbf{h} \in \mathbb{R}^{T}
\end{align}

A word-wise sigmoid is then applied to element \(h_t\) as similar to computing the selection probabilities:
\begin{align}
\pi_\theta(\mathbf{x}, \{\mathcal{K}^{(k)}(\mathbf{x})\})_t = \text{Sigmoid}(h_t)
\end{align}
We then sample a binary mask \(\mathbf{m} \sim \texttt{Bern}(\pi_\theta(\mathbf{x}, \{\mathcal{K}^{(k)}(\mathbf{x})\}))\) to obtain the selected subset. The reduced input \(\tilde{\mathbf{x}}\) is formed by replacing unselected words with a neutral placeholder token.

Since the Graph-SST2 dataset \cite{yuan_graphsst2_IEEE2022} readily provides semantic graph structures, we utilized it to evaluate our method with both syntactic and semantic graphs; these results are presented in Table \ref{tab:graph_sst2_all}.

\textbf{Graph-SST2} \cite{yuan_graphsst2_IEEE2022}: A graph-structured version of the SST2 dataset, where each sentence is augmented with a dependency parse graph constructed via the Biaffine parser~\cite{dozat_biaffine_ICLR2017}. Each node corresponds to a word with contextualized BERT embeddings~\cite{devlin_Bert_NAACL2019} as features, and the task is binary sentiment classification. This dataset enables evaluation of structure-aware selectors under semantically rich graph input.

\subsection{Benchmarks} \label{appendix: benchmarks}
Throughout our experiments, we compare \textbf{GLASS} against both memoryless explanation methods --- such as LIME~\cite{ribeiro_LIME_KDD2016}, KernelSHAP (which we denote as SHAP)~\cite{lundberg_shap_NeurIPS2017}, and Integrated Gradients (IG)~\cite{sundararajan_ig_ICML2017} --- and amortized approaches, including L2X~\cite{chen_l2x_ICML2018}, CXplain~\cite{schwab_cxplain_NeurIPS2019} and LTX~\cite{barkan_LTX_ICDM2023}. We focus on these widely recognized baselines in the main manuscript to ensure clarity and reproducibility under strict black-box conditions. Additional comparisons with the recent state-of-the-art methods~\cite{barkan_LTX_ICDM2023, schwab_cxplain_NeurIPS2019} are provided in this supplementary material. 
For completeness, we also include INVASE~\cite{yoon_invase_ICLR2018}, a representative earlier work on instance-wise feature selection, adapted for NLP tasks. 
In addition, we investigate two representative approaches for feature selection: the Hard Concrete (HC) distribution~\cite{louizos_hardconcrete_ICLR2018}, and the Straight-Through Estimator (STE)~\cite{bengio_STE_2013}. To analyze soft selection via continuous relaxation, we adopt HC, which enables training through soft gates. In contrast, STE allows learning discrete gates in settings where gradient access is available. We implement both methods within our model structure, replacing the RL-based selection module, in order to isolate and compare the effect of soft versus discrete selection under different optimization strategies.

For comparative analysis, we include baselines that typically operate under \textit{weak black-box} or \textit{white-box} conditions. To enable this, we grant these methods \textit{semi-oracle} or \textit{oracle} access to the target black-box DLM internals, specifically by allowing them to utilize token embeddings (for LIME, SHAP, and CXplain) or gradients (for IG, LTX, L2X, INVASE, Hard Concrete, and GLASS with STE) that are generally inaccessible in a practical black-box setting.
More details about each method are provided below:

\noindent 
\textbf{Memoryless Explanation Methods.}~~
\begin{itemize}[leftmargin=*, topsep=0pt, partopsep=0pt, itemsep=0pt]
    \item \textbf{LIME}~\cite{ribeiro_LIME_KDD2016}: 
    A perturbation-based method that fits a local linear model to approximate the decision boundary around the input by masking subsets of features. For text inputs, words are masked by replacing their constituent tokens with \texttt{[PAD]}, and cosine similarity weights the perturbed instances. We adhere to the standard \texttt{Thermostat} library implementation~\cite{feldhus_thermostat_EMNLP2021}, using a 0.2 per-token masking probability and 256 samples per explanation.

    \item \textbf{SHAP}~\cite{lundberg_shap_NeurIPS2017}: 
    A game-theoretic attribution method that estimates Shapley values by computing the marginal contribution of each feature across all possible feature subsets. KernelSHAP is used to approximate such marginal contribution in a model-agnostic setting. For text inputs, tokens are masked by replacing their constituent tokens with \texttt{[PAD]}. We use the default 100 background samples via \texttt{Thermostat} for computationally tractable approximation. 

    \item \textbf{Integrated Gradients (IG)}~\cite{sundararajan_ig_ICML2017}: 
    A notable feature attribution technique involves integrating gradients along a linear path connecting a baseline input to the target input. This process inherently requires access to the target model's internal gradients across these input variations, thereby restricting its applicability to white-box settings. 
    We use 25 interpolation steps and a sequence of the neutral tokens (i.e., \texttt{[PAD]}) as the baseline input, consistent with the \texttt{Thermostat} library implementation~\cite{feldhus_thermostat_EMNLP2021}. 
\end{itemize}
For methods such as LIME, SHAP, and IG, which provide importance scores for individual input features, we measured their discriminative performance based on these scores. To achieve this, we first defined a fixed selection rate (e.g., 10\%, 30\%) from the entire population of words across all samples. We then selected the top N\% of words based on their calculated importance scores from this total population. For evaluation, only these selected words were retained, while the remaining words were masked using the model's attention mask. This masking technique effectively reduced the influence of unselected words on the model's output, allowing us to assess the discriminative power of the chosen subsets.

\noindent 
\textbf{Amortization Explanation Methods.}~~
To ensure a fair comparison, we initialize the word representations in all amortized baselines (\cite{barkan_LTX_ICDM2023,chen_l2x_ICML2018, yoon_invase_ICLR2018}) using the same pretrained DLM encoders (i.e., the word embedding model) as used in \textbf{GLASS}. This ensures that performance differences arise solely from the selector architectures and training strategies, not the underlying representations.

We further unify the architectural and training setup across all amortized methods by  (i) adapting each method to a \emph{label-centric} formulation to improve the discriminative power and (ii) replacing surrogate models by providing \textit{oracle access} to gradients from the black-box predictor. All selectors are implemented using a two-layer MLP applied to frozen word embeddings. We refer to these enhanced versions as L2X and treat LTX as a sparsity-regularized variant of L2X.

To enhance training efficiency for interpreting large, frozen black-box predictors, our method focuses on training only compact GNN or MLP modules as word selectors, crucially avoiding the construction of a surrogate model to approximate the predictor.
This contrasts with a common strategy for enabling gradient-based optimization in discrete selection tasks: the use of continuous relaxations (e.g., Gumbel-Softmax \cite{paulus_Gumbel_NeurIPS2020} or the Hard Concrete~\cite{louizos_hardconcrete_ICLR2018} tricks). Such relaxation methods produce soft mask vectors that average token embeddings, effectively generating synthetic, out-of-vocabulary word representations. These hybrid inputs are \textit{out-of-distribution (OOD)} for our target black-box DLM that is not exposed to them during its training, which can result in degraded or unstable outputs. 
Critically, when such relaxations are used, any noisy or biased gradients arising from the black-box DLM's OOD responses are then backpropagated directly to the selector. A compact selector, due to its limited capacity, would be particularly vulnerable to these distorted signals, risking performance degradation.
While the Straight-Through Estimator (STE) \cite{bengio_STE_2013} allows optimization with hard masks and preserves discrete word selection, it introduces its own gradient bias, which can also lead to suboptimal training.

In contrast, policy gradient methods preserve the discrete nature of word selection and provide unbiased gradient estimates, offering a more robust and principled optimization framework. We present experimental results comparing variants of our methods using Hard Concrete and STE against our proposed method, \textbf{GLASS} (trained using policy gradients).

\begin{itemize}[leftmargin=*, topsep=0pt, partopsep=0pt, itemsep=0pt]
\item \textbf{CXplain}\cite{schwab_cxplain_NeurIPS2019}:
While CXPlain offers a general framework for model-agnostic feature importance estimation, its direct application at the word level introduces practical and conceptual challenges, particularly in natural language processing (NLP) tasks. In the original implementation, CXPlain computes feature importance by individually removing each feature from the input and measuring the resulting change in model performance. This naive leave-one-feature-out approach scales linearly with input sequence length, rendering it computationally inefficient for long documents or large datasets.

More importantly, in typically long text inputs, the contribution of each individual word tends to be marginal and often context-dependent. As such, removing a single word may result in a negligible change in the model’s output, leading to unreliable or noisy importance scores. Evaluating all possible word combinations or higher-order interactions would be theoretically more informative, but it is computationally intractable in practice due to the exponential number of combinations. For our benchmark, we implemented CXPlain following the baseline approach by masking one word at a time by replacing its constituent tokens with \texttt{[PAD]}. 

\item \textbf{L2X}~\cite{chen_l2x_ICML2018}: 
L2X selects informative subsets by maximizing a variational lower bound on mutual information between the selected feature subset and the target black-box model. In our setup, we minimize cross-entropy loss on selected words (i.e., the corresponding tokens) using gradients from the black-box model. In the main paper, we have used a simplified version of L2X, implemented as a two-layer MLP selector on top of frozen DLM embeddings, to enable minimal and efficient comparison under space constraints. 
For baseline comparison, we used a sampling number $k=10$. However, for the Movies dataset, this value was insufficient for adequate exploration due to its longer average sentence lengths. Therefore, to ensure training stability, we increased the parameter to $k=100$ exclusively for this dataset.

\item \textbf{LTX}~\cite{barkan_LTX_ICDM2023}: 
A black-box explainer that learns soft selection masks via differentiable subset sampling. Although the original formulation includes inversion and smoothness regularization terms (i.e., \textit{L\textsubscript{inv}}, \textit{L\textsubscript{smooth}} in \cite{barkan_LTX_ICDM2023}), these components are omitted in our experiments following the original implementation. LTX is already aligned with our experimental setting, as it minimizes cross-entropy loss with respect to the true label using gradients directly from the black-box model. As such, it inherently assumes oracle access. Conceptually, LTX can be viewed as a sparsity-regularized variant of \textbf{OracleL2X}, where the selection target is fixed to an all-zero mask. We set $k = 100$ for the Movies dataset.

Originally designed for vision tasks, LTX does not account for the discrete nature of textual inputs and employs the Gumbel-Softmax trick without additional constraints. In its original setup, the selector is initialized by copying the weights of the frozen black-box model, and an additional MLP layer for the selection is trained. For a fair and efficient comparison under our setting, we instead initialize the word embeddings using pre-trained DLM embeddings (i.e., the word embedding network) and only train a small selector consisting of a two-layer MLP.

\item \textbf{INVASE}~\cite{yoon_invase_ICLR2018}: \label{appendix: invase_modified}
The original INVASE framework jointly trains three components: a selector (actor), a predictor on selected features (critic), and a baseline model that uses the full input. However, this setup does not directly explain a fixed target model, since both the critic and baseline are learned from scratch.

To adapt INVASE for our setting—where the goal is to explain an existing, fixed black-box model—we modify the framework so that the target black-box DLM itself is used for both the critic and the baseline. Specifically, we feed selected features into the target model to obtain the critic output, and use the same model on the full input as the baseline.
Furthermore, as mentioned in the main text, we use a pre-trained DLM as the embedding model and process embeddings with shared MLPs, an architectural choice adapted from the model to ensure fair comparison, to ensure that the selector handles variable-length inputs robustly.

\end{itemize}

\noindent\textbf{Cost Rating Criteria for Table~\ref{table:summary}.}~~ \textit{Training Cost} refers to the cost of training a \emph{reusable} model prior to inference. Methods that require no pre-trained explainer (IG, LRP, DeepLIFT, PDA, KernelSHAP, LIME) are rated \textit{None}, even though some (e.g., LIME) fit a local surrogate at inference time --- this cost is instead reflected in \textit{Inference Cost}. Methods training only a lightweight selector without a surrogate (LTX, Ours) are rated \textit{Low}. Methods jointly training a selector and surrogate predictor (INVASE, L2X) are rated \textit{Moderate}. CXPlain is rated \textit{High} as it requires exhaustive feature ablation to construct supervised training labels.

\textit{Inference Cost} reflects the computational burden at test time. Perturbation-based methods (KernelSHAP, LIME) are rated \textit{High} as they require numerous model queries per input to estimate feature importance. IG is similarly rated \textit{High} due to the repeated gradient computations along the integration path. In contrast, LRP and DeepLIFT are rated \textit{Low} as they compute attributions in a single forward/backward pass. All amortized methods including \proposed~are rated \textit{Low}, as the trained selector produces explanations in a single pass at test time without any additional model queries (see Table~\ref{tab:efficiency}).

\subsection{Ablation Study} \label{appendix: ablation_study}

\begin{table}[t]
\centering
\caption{\small Performance comparison across different neutral placeholder strategies on the Movies dataset. The same word selector is used across all variants.}
\label{tab:placeholder_ablation}
\begin{tabular}{lccc}
\toprule
\textbf{Masking Strategy} & \textbf{ACC} & \textbf{AUROC} & \textbf{AUPRC} \\
\midrule
\rowcolor[HTML]{FDEBD0}Replace with \; \textbf{``the''}    & \textbf{0.900}& \textbf{0.966}& \textbf{0.968}\\ 
 Replace with \; \textbf{``,''} & 0.779 & 0.856 & 0.842 \\
Replace with \; \textbf{``\_''}               & 0.573 & 0.628	 & 0.569 \\
 Replace with \; \textbf{``blank''  }& 0.638 & 0.759 & 0.750\\
Replace with \; \textbf{[UNK]}               & 0.749 & 0.895	 & 0.894 \\
Replace with \; \textbf{[PAD]}               & 0.739 & 0.889	 & 0.893 \\
Replace with \; \textbf{[MASK]}               & 0.709 & 0.833	 & 0.800 \\
 Remove word                      & 0.754 & 0.814	 & 0.743 \\
 \midrule
Oracle attention masking          & 0.849 & 0.914	 & 0.901 \\
\bottomrule
\end{tabular}
\end{table}

\noindent 
\textbf{Effect of Masking Strategy.}~~
Table~\ref{tab:placeholder_ablation} compares the performance of different masking strategies applied to unselected words on the Movies dataset. All experiments use the same word selector to isolate the impact of the placeholder choice during explanation.

Among the strategies, replacing unselected words with the word “the” results in the best overall performance, likely due to its high frequency and syntactic neutrality within English sentences. Oracle masking, which manipulates the attention mask directly, also performs strongly, reflecting the benefit of preserving word positions while hiding content. In contrast, using the visually neutral or structurally ambiguous token, such as underscores, leads to substantial performance degradation. Special tokens like \texttt{[PAD]}, \texttt{[MASK]}, and \texttt{[UNK]} yield intermediate results, with \texttt{[PAD]} and \texttt{[UNK]} outperforming \texttt{[MASK]}, suggesting that the model may treat masking-related tokens differently depending on their pretraining context.

During training, we experimented with different masking tokens and found an intriguing pattern. Validation revealed that \textsc{STE}, \textsc{L2X}, and \textsc{LTX} achieve their highest scores when the placeholder word is \textbf{``blank''}, whereas the Hard-Concrete models—\textsc{MLP(HC)} and \textsc{GCN(HC)}—perform best when the placeholder is the high-frequency article \textbf{``the''}. Interestingly, substituting the placeholder with ``blank'' does not help our \textbf{GLASS} model—in fact, it lowers every metric compared to using ``the''. This contrast suggests that, beyond function words (articles, conjunctions, etc.), there exist content-bearing tokens whose embeddings align more favorably with specific selector–predictor pairs. The ``blank'' vector appears to strike a balance: it is neither too frequent (thus avoiding bias toward common co-occurrences) nor entirely out-of-distribution (remaining within the pre-trained embedding manifold), providing a ``neutral yet well-formed'' replacement that some soft or STE-based methods can exploit—though not universally.

Overall, these findings highlight that the choice of placeholder is not trivial: certain tokens better preserve sentence structure or align more naturally with the model’s learned expectations, which in turn affects the faithfulness of the resulting explanations. Based on validation performance, we replace the masked word with ‘the’ in the Movies, HateXplain dataset and remove the word in the IMDb, Graph-SST2, XSum dataset.

\begin{table*}[t] 
\centering
\caption{\small \textbf{Movies}: ACC, AUROC, and AUPRC scores for each graph type across selection rates.}
\label{tab:graph_type_comparison_full}
\resizebox{\textwidth}{!}{%
\begin{tabular}{ll|ccc|ccc|ccc}
    \toprule
    \multicolumn{2}{c|}{\multirow{2}{*}{\textbf{Graph Type}}} & \multicolumn{3}{c|}{\textbf{0.05}} & \multicolumn{3}{c|}{\textbf{0.10}} & \multicolumn{3}{c}{\textbf{0.15}} \\
    \cmidrule(lr){3-5} \cmidrule(lr){6-8} \cmidrule(l){9-11}
    \multicolumn{2}{c|}{} & ACC & AUROC & AUPRC & ACC & AUROC & AUPRC & ACC & AUROC & AUPRC \\
    \midrule
    & MLP   
    &   0.567 ± 0.029   &   0.761 ± 0.031   &   0.767 ± 0.030   
    &  0.733 ± 0.015    &  0.805 ± 0.016	    &   0.797 ± 0.025   
    &   0.720 ± 0.027  &   0.816 ± 0.035   &   0.808 ± 0.038	   \\
   \rowcolor[HTML]{F5F5F5 } & MLP (w/ LP)     
    &    0.578 ± 0.019	  &   0.789 ± 0.021   &   0.788 ± 0.018   
    &   0.733 ± 0.021  &  0.808 ± 0.019  &  0.797 ± 0.023	
    &  0.686 ± 0.01   &    0.782 ± 0.011	  &    0.779 ± 0.014	  \\
    \rowcolor[HTML]{FFF5E6}& GCN (w/ Fully)   
    &   0.547 ± 0.008   &    0.543 ± 0.007	  &   0.595 ± 0.015   
    &   0.577 ± 0.011   &  0.572 ± 0.011	   &   0.642 ± 0.023	   
    &   0.616 ± 0.014   &   0.611 ± 0.016	   &   0.693 ± 0.026   \\
   \rowcolor[HTML]{FFF5E6} & GCN (w/ Serial)   
    &   0.574 ± 0.041   &    0.613 ± 0.061	  &   0.661 ± 0.048   
    &   0.663 ± 0.071   &  0.765 ± 0.125	   &   0.785 ± 0.132	   
    &   0.750 ± 0.089   &   0.611 ± 0.136	   &   0.810 ± 0.145   \\
    \rowcolor[HTML]{FDEBD0} & GCN (w/ Syn) (\textbf{Ours})     
    &   \textbf{0.639 ± 0.029}   &   \textbf{0.900 ± 0.010}	   &  \textbf{0.927 ± 0.008}   
    &   \textbf{0.829 ± 0.017}   &   \textbf{0.954 ± 0.012}   &   \textbf{0.962 ± 0.009}   
    &   \textbf{0.895 ± 0.006}	  &   \textbf{0.951 ± 0.006}   &    \textbf{0.954 ± 0.006}  \\
    \bottomrule
    \midrule
    \multicolumn{2}{l|}{Oracle}
    & \multicolumn{3}{c}{}  
    & \multicolumn{1}{c}{0.859} 
    & \multicolumn{1}{c}{0.942} 
    & \multicolumn{1}{c}{0.944} 
    & \multicolumn{3}{c}{}  \\
    \bottomrule
\end{tabular}
}
\end{table*}

\begin{table*}[t] 
\centering\caption{\small  \textbf{Graph SST-2}: ACC, AUROC, and AUPRC scores for different graph types and selection rates.}
\label{tab:graph_sst2_all}

\resizebox{\linewidth}{!}{%
\begin{tabular}{l|ccc|ccc|ccc}
\toprule
\textbf{Graph Type}& \multicolumn{3}{c|}{\textbf{0.05}} 
& \multicolumn{3}{c|}{\textbf{0.10}} 
& \multicolumn{3}{c}{\textbf{0.15}} \\
\midrule
& ACC & AUROC & AUPRC 
  & ACC & AUROC & AUPRC 
  & ACC & AUROC & AUPRC \\
\midrule

MLP              
& 0.623 ± 0.020   &  0.683 ± 0.023	  &  0.686 ± 0.013  
&  0.702 ± 0.036    &  0.781 ± 0.036    &  0.770 ± 0.029		
&  0.737 ± 0.038    &  0.822 ± 0.032   &  0.814 ± 0.027	    \\
\rowcolor[HTML]{F5F5F5 }MLP (w/ Syn LP)&  0.675 ± 0.030    &   0.738 ± 0.024   &   0.682 ± 0.039   
&  0.724 ± 0.038    &   0.793 ± 0.033   &  0.760 ± 0.0293         
&  0.746 ± 0.033    &   0.820 ± 0.024    &  0.798 ± 0.018	   \\
\rowcolor[HTML]{F5F5F5 }MLP (w/ Sem LP)&  0.668 ± 0.019    &  0.724 ± 0.021    &  0.668 ± 0.036   
& 0.724 ± 0.036	     &  0.791 ± 0.019	     &  0.743 ± 0.010 
&  0.753 ± 0.035    &  0.826 ± 0.017	 & 0.790 ± 0.009     \\
\rowcolor[HTML]{F5F5F5 }MLP (w/ Syn \& Sem LP)&   0.655 ± 0.020   &  0.715 ± 0.039    &   0.690 ± 0.041
&  0.713 ± 0.027    &   0.786 ± 0.020	   &   0.761 ± 0.019
&  0.733 ± 0.028    &   0.815 ± 0.020    &   0.805 ± 0.018    \\
\rowcolor[HTML]{FDEBD0}GraphSAGE (w/ Syn)&  0.725 ± 0.019	    &   0.824 ± 0.008   & 0.810 ± 0.010   
& 0.767 ± 0.031    &  0.864 ± 0.022	    & 0.860 ± 0.022    
&  0.790 ± 0.024    &  0.880 ± 0.017	    &   0.880 ± 0.015    \\
\rowcolor[HTML]{FDEBD0}GraphSAGE (w/ Sem)&  \textbf{0.738 ± 0.015}& 0.832 ± 0.007    &  0.819 ± 0.010   
&  \textbf{0.788} ± 0.021&  \textbf{0.884 ± 0.015}& \textbf{0.880 ± 0.013}& 0.798 ± 0.016    &  \textbf{0.893 ± 0.009}&  0.891 ± 0.011    \\
\rowcolor[HTML]{FDEBD0}GraphSAGE (w/ Syn \& Sem)&  0.735 ± 0.018    &  \textbf{0.834 ± 0.011}& \textbf{0.826 ± 0.010}& 0.784 ± 0.025 &   0.881 ± 0.020	   &   0.879 ± 0.015	  
&   \textbf{0.806 ± 0.025}&   0.806 ± 0.025   &  \textbf{0.900 ± 0.017}\\
\bottomrule
\midrule
Oracle& \multicolumn{3}{c}{}  
& \multicolumn{1}{c}{0.842} 
& \multicolumn{1}{c}{0.926} 
& \multicolumn{1}{c}{0.928} 
& \multicolumn{3}{c}{}  \\
\bottomrule
\end{tabular}%
}
\end{table*}

\noindent 
\textbf{Graph Type and Selector Architecture.}~~
We conduct ablation studies to evaluate the impact of graph type and selector architecture under varying selection rates (5\%, 10\%, and 15\%). As shown in Table~\ref{tab:graph_type_comparison_full}, we compare MLP-based selectors with and without Laplacian regularization (LP), as well as GCN-based variants using either fully connected graphs, syntactic dependency graphs, or serial graphs, which simply connect words in their sentence order. These results demonstrate that incorporating syntactic structure via GCNs not only enhances the interpretability of selected words, as shown in the main paper, but also improves their discriminative quality for the downstream task. Notably, the GCN with syntactic graphs achieves strong performance across all selection rates, indicating that structural linguistic information is beneficial even under budget constraints. Moreover, the performance gap between syntactic and fully connected graphs confirms that the gains are not simply due to the GCN architecture itself, but rather arise from effective use of the language structure of the input.

We further analyze the impact of incorporating syntactic and semantic edge types in the Graph-SST2 dataset (Table~\ref{tab:graph_sst2_all}).  The semantic graphs, provided in the dataset, tend to be densely connected and potentially noisy. As GCNs often suffer from oversmoothing, reducing their discriminative capacity, we adopt GraphSAGE~\cite{hamilton_GraphSAGE_NeurIPS2017}, which performs neighbor sampling and localized aggregation, enabling robust learning on large or densely connected graphs. Graph-based models consistently outperform MLP-based selectors across all evaluation metrics and selection rates, highlighting the benefit of leveraging structural information. However, while both syntactic and semantic edges contribute positively when used individually, combining them does not lead to substantial performance gains beyond using either one alone. This trend is consistent across both MLP and GraphSAGE architectures, suggesting that the redundancy or overlap between syntactic and semantic signals may limit the additive benefit of combining them.

\begin{table*}[t]
\centering
\caption{\small \textbf{Graph-SST2}: ACC, AUROC, and AUPRC scores across selection rates for different encoder-selector combinations.
}
\label{tab:encoder_selector_comparison}
\resizebox{\linewidth}{!}{%
    \begin{tabular}{cc|ccc|ccc|ccc}
    \toprule
    \textbf{$f$}& \textbf{$g$}& \multicolumn{3}{c|}{\textbf{0.05}} 
    & \multicolumn{3}{c|}{\textbf{0.10}} 
    & \multicolumn{3}{c}{\textbf{0.15}} \\
    \cmidrule(lr){3-5} \cmidrule(lr){6-8} \cmidrule(lr){9-11}
    & & ACC & AUROC & AUPRC 
      & ACC & AUROC & AUPRC 
      & ACC & AUROC & AUPRC \\
    \midrule
    BERT        & & 0.725 ± 0.019 & 0.824 ± 0.008	 & 0.810 ± 0.010
    & 0.767 ± 0.031 & 0.864 ± 0.022	 & 0.860 ± 0.022	
    & 0.790 ± 0.024 & 0.880 ± 0.017	 & 0.880 ± 0.015	 \\
    DeBERTa-v3  & DeBERTa-v3\textsuperscript{*}    
    & 0.771 ± 0.007 & 0.868 ± 0.009 & 0.854 ± 0.016
    & 0.840 ± 0.006 & 0.930 ± 0.006 & 0.925 ± 0.009	
    & 0.855 ± 0.007 & 0.936 ± 0.005	 & 0.934 ± 0.007 \\
    RoBERTa     & & 0.753 ± 0.017 & 0.865 ± 0.007 & 0.861 ± 0.004
    & 0.823 ± 0.003 & 0.920 ± 0.004	 & 0.917 ± 0.005
    & 0.840 ± 0.006 & 0.924 ± 0.007	 & 0.924 ± 0.005	 \\
\midrule
    BERT      & & 0.779 ± 0.013	& 0.872 ± 0.016	& 0.876 ± 0.018
    & 0.827 ± 0.006	& 0.911 ± 0.002	& 0.913 ± 0.003
    & 0.847 ± 0.003	& 0.922 ± 0.001	& 0.926 ± 0.001 \\
    DeBERTa-v3      & GPT-2\textsuperscript{*}  
    & 0.852 ± 0.005& 0.925 ± 0.006 & 0.922 ± 0.008
    & 0.887 ± 0.003	& 0.950 ± 0.003 & 0.950 ± 0.004
    & 0.886 ± 0.001	& 0.952 ± 0.002	& 0.954 ± 0.002 \\
    RoBERTa      & &0.814 ± 0.017	& 0.912 ± 0.016	& 0.915 ± 0.017
    &0.869 ± 0.007	& 0.940 ± 0.004	& 0.941 ± 0.005
    &0.877 ± 0.004	& 0.942 ± 0.002	& 0.945 ± 0.002 \\
    \bottomrule
    \midrule
    \multirow{2}{*}{Oracle}& DeBERTa-v3& \multicolumn{3}{c}{} %
    & \multicolumn{1}{c}{0.842}       %
    & \multicolumn{1}{c}{0.926}
    & \multicolumn{1}{c}{0.928}
    & \multicolumn{3}{c}{}  \\
    
    & GPT-2
    & \multicolumn{3}{c}{} %
    & \multicolumn{1}{c}{0.819}       %
    & \multicolumn{1}{c}{0.907}
    & \multicolumn{1}{c}{0.913}
    & \multicolumn{3}{c}{}  \\
    \bottomrule
\end{tabular}
}
\begin{flushleft}
\textit{\textsuperscript{*}The predictor is task-specific and fine-tuned on the full training set.}
\end{flushleft}
\end{table*}

\begin{table*}[t] 
\centering
\caption{\small \textbf{Movies}:  ACC, AUROC, and AUPRC scores for different DeBERTa encoder sizes.}
\label{tab:deberta_size_comparison}
\resizebox{\linewidth}{!}{%
\begin{tabular}{l|ccc|ccc|ccc}
\toprule
\textbf{$f$}& \multicolumn{3}{c|}{\textbf{0.05}} 
& \multicolumn{3}{c|}{\textbf{0.10}} 
& \multicolumn{3}{c}{\textbf{0.15}} \\
\cmidrule(lr){2-4} \cmidrule(lr){5-7} \cmidrule(lr){8-10}
DeBERTa-v3 & ACC & AUROC & AUPRC 
& ACC & AUROC & AUPRC 
& ACC & AUROC & AUPRC \\
\midrule
\rowcolor[HTML]{FFF5E6}Small& 0.580 ± 0.032 & 0.583 ± 0.084 & 0.667 ± 0.089 
& 0.735 ± 0.017 & 0.818 ± 0.033 & 0.847 ± 0.025 
& 0.756 ± 0.023 & 0.835 ± 0.022 & 0.840 ± 0.027 \\
\rowcolor[HTML]{FDEBD0}Base& \textbf{0.639 ± 0.029}& \textbf{0.900 ± 0.010}& \textbf{0.927 ± 0.008}& \textbf{0.829 ± 0.017}& \textbf{0.954 ± 0.012}& \textbf{0.962 ± 0.009}& \textbf{0.895 ± 0.006}& \textbf{0.951 ± 0.006}& \textbf{0.954 ± 0.006}\\
\rowcolor[HTML]{FFF5E6}Large& 0.592 ± 0.019 & 0.633 ± 0.032 & 0.721 ± 0.030 
& 0.782 ± 0.033 & 0.866 ± 0.020 & 0.886 ± 0.023
& 0.797 ± 0.032 & 0.876 ± 0.026 & 0.880 ± 0.029 \\
\bottomrule
\midrule
\multicolumn{1}{l|}{\textbf{Oracle}}
& \multicolumn{3}{c}{}  
& \multicolumn{1}{c}{0.824} 
& \multicolumn{1}{c}{0.936} 
& \multicolumn{1}{c}{0.940} 
& \multicolumn{3}{c}{}  \\
\bottomrule
\end{tabular}
}
\end{table*}

\noindent 
\textbf{Word Embedding Model Architecture and Scale.}~~
We further analyze the effect of word embedding model choice on the selector performance. In Table~\ref{tab:encoder_selector_comparison}, we evaluate the performance of various combinations of selector embedding models and target black-box predictors. We initialize selectors with different word embedding models $f: \mathcal{X}^{T} \rightarrow \mathbb{R}^{T\times d}$, namely BERT, RoBERTa, and DeBERTa-v3, and test them against predictors based on both encoder (DeBERTa-v3) and decoder (GPT-2) architectures. This expanded setup allows us to analyze not only the effect of the selector's representation space but also its compatibility across different predictor architectures.

In this analysis, we restrict the selector's embedding backbone to encoder-based models, as word selection fundamentally requires token representations that reflect the full-sentence context. Encoder-based models employ bidirectional attention mechanisms, enabling each word representation to incorporate context from both preceding and following words. This capability is particularly critical for our embedding model, as determining the importance of a word requires understanding its role within the complete sentence structure. In contrast, decoder-based models (e.g., GPT-2, LLaMA) utilize causal (unidirectional) attention, restricting each token to attend only to preceding ones. This architectural constraint forces the reliance on token representations that reflect only the preceding context rather than the entire input text, thereby limiting their capacity to capture the full contextual information necessary for effective word selection.

Table~\ref{tab:deberta_size_comparison} reports the performance of selectors, whose word embedding is initialized with different embedding model sizes --- \textit{small}, \textit{base}, and \textit{large} --- within the DeBERTa-v3 family. Across all selection rates, the \textit{base} encoder achieves the best performance. Transitioning from \textit{small} to \textit{base} leads to substantial gains, while further increasing to \textit{large} offers no additional benefit and may even lead to diminished performance.

One possible reason for the strong performance of the \textit{base} word embedding model $f$ is that it possesses a similar model capacity to the black-box DLM, which is also based on the DeBERTa-v3 \textit{base}. Although the predictor is fine-tuned on the task and thus differs in parameters and learned representations, using a version of $f$ with comparable capacity may help extract word embeddings that align more closely with the predictor’s decision boundaries. These results suggest that larger embedding models do not always yield better performance; rather, aligning the capacity of the embedding model $f$ with that of the predictor may be more critical. This highlights that representational compatibility---not model size alone---plays a key role in identifying informative input words.

\begin{table*}[t]
\centering
\caption{\small \textbf{Movies}: Performance (Accuracy, AUROC, and AUPRC) across selection rates using different encoder output layers.}
\label{tab:encoder_layer_depth_comparison}
\resizebox{\linewidth}{!}{%
\begin{tabular}{l|ccc|ccc|ccc}
\toprule
\textbf{Encoder Layer (Depth)} 
& \multicolumn{3}{c|}{\textbf{0.05}} 
& \multicolumn{3}{c|}{\textbf{0.10}} 
& \multicolumn{3}{c}{\textbf{0.15}} \\
\cmidrule(lr){2-4} \cmidrule(lr){5-7} \cmidrule(lr){8-10}
& ACC & AUROC & AUPRC 
& ACC & AUROC & AUPRC 
& ACC & AUROC & AUPRC \\
\midrule
\rowcolor[HTML]{F5F5F5 }Non-contextual Embedding& 0.512 ± 0.015 & 0.501 ± 0.040 & 0.496 ± 0.038 
& 0.524 ± 0.008 & 0.541 ± 0.024 & 0.519 ± 0.018 
& 0.537 ± 0.011 & 0.588 ± 0.017 & 0.566 ± 0.027	 \\
\rowcolor[HTML]{F5F5F5 }(+ Position Encoding)& 0.516 ± 0.016 & 0.560 ± 0.022 & 0.543 ± 0.044 
& 0.527 ± 0.020 & 0.561 ± 0.027 & 0.554 ± 0.026 
& 0.560 ± 0.025 & 0.590 ± 0.019 & 0.562 ± 0.027 \\
\rowcolor[HTML]{FFF5E6}Layer 1 (First)& 0.546 ± 0.010	 & 0.610 ± 0.044 & 0.583 ± 0.042 
& 0.521 ± 0.011 & 0.489 ± 0.045 & 0.478 ± 0.033 
& 0.528 ± 0.021 & 0.484 ± 0.048 & 0.482 ± 0.032 \\
\rowcolor[HTML]{FFF5E6}Layer 5 (Middle)& 0.612 ± 0.038 & 0.717 ± 0.040 & 0.670 ± 0.053 
& 0.530 ± 0.014 & 0.416 ± 0.042 & 0.428 ± 0.021 
& 0.522 ± 0.008 & 0.360 ± 0.034 & 0.394 ± 0.017 \\
\rowcolor[HTML]{FDEBD0}Layer 10 (Used)& \textbf{0.640 ± 0.029}& \textbf{0.895 ± 0.013}& \textbf{0.925 ± 0.009}& \textbf{0.826 ± 0.018}& \textbf{0.953 ± 0.013}& \textbf{0.961 ± 0.010}& \textbf{0.897 ± 0.007}& \textbf{0.954 ± 0.008}& \textbf{0.955 ± 0.008}\\
\rowcolor[HTML]{FFF5E6}Layer 12 (Last)& 0.580 ± 0.039 & 0.694 ± 0.033 & 0.664 ± 0.036 
& 0.544 ± 0.015 & 0.526 ± 0.046 & 0.502 ± 0.030 
& 0.558 ± 0.006 & 0.475 ± 0.023 & 0.447 ± 0.013 \\
\bottomrule
\midrule
\multicolumn{1}{l|}{\textbf{Oracle}}
& \multicolumn{3}{c}{}  
& \multicolumn{1}{c}{0.824} 
& \multicolumn{1}{c}{0.936} 
& \multicolumn{1}{c}{0.940} 
& \multicolumn{3}{c}{}  \\
\bottomrule
\end{tabular}
}
\end{table*}

\noindent 
\textbf{Effect of Encoder Layer Depth.}~~
To assess how the depth of the encoder affects the quality of feature selection, we conduct an ablation study using embeddings extracted from various layers of the pre-trained DeBERTa-v3 \textit{base} model. As shown in Table~\ref{tab:encoder_layer_depth_comparison}, intermediate layers, particularly Layer 10, consistently yield the best performance across all evaluation metrics and selection rates.

While it may seem intuitive to use the final encoder layer for downstream tasks, we find that it performs significantly worse than intermediate layers. We conjecture that this is due to the final layer being heavily specialized for the pretraining objectives --- namely, masked language modeling (MLM) and replaced token detection (RTD). As our encoder is not fine-tuned during training, the representations from the final layer may not transfer well to the word selection task, leading to suboptimal results. In contrast, embeddings from the 10th layer appear to strike a better balance between general semantic information and task-agnostic structure, making them more suitable for word selection.

\begin{figure*}[!t]
\centering
\includegraphics[trim={0cm 0cm 0cm 0cm}, clip, width=0.75\linewidth]{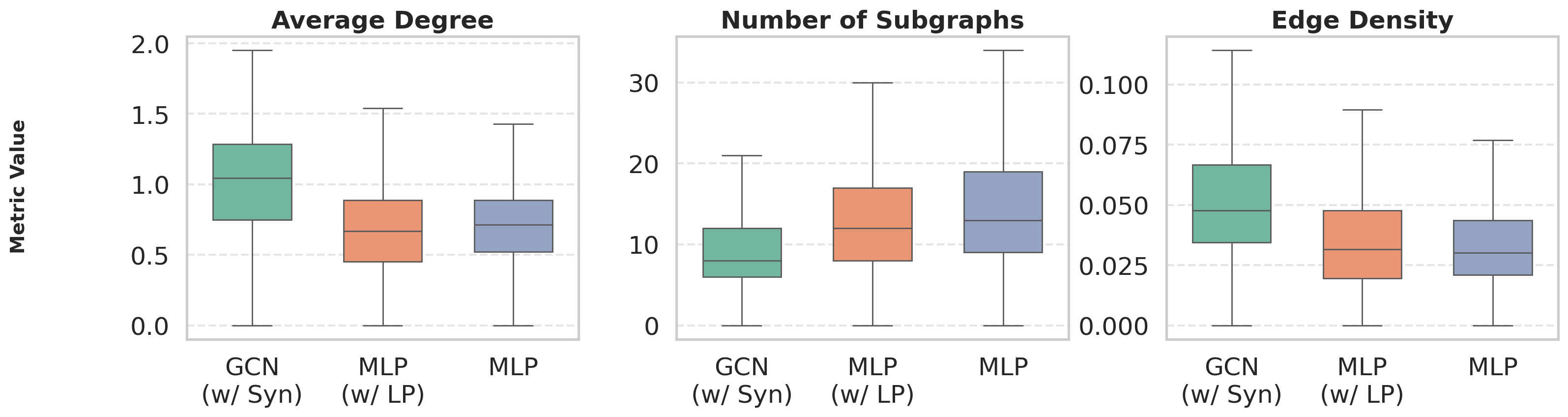}
\caption{\small Comparison of linguistic coherence with respect to (left) average node degree (middle) the number of subgraphs (right) edge density in the IMDb dataset.}
\label{fig:IMDb_structure_coherence}
\end{figure*}

\begin{figure*}[!t]
\centering
\includegraphics[trim={0cm 0cm 0cm 0cm}, clip, width=0.75\linewidth]{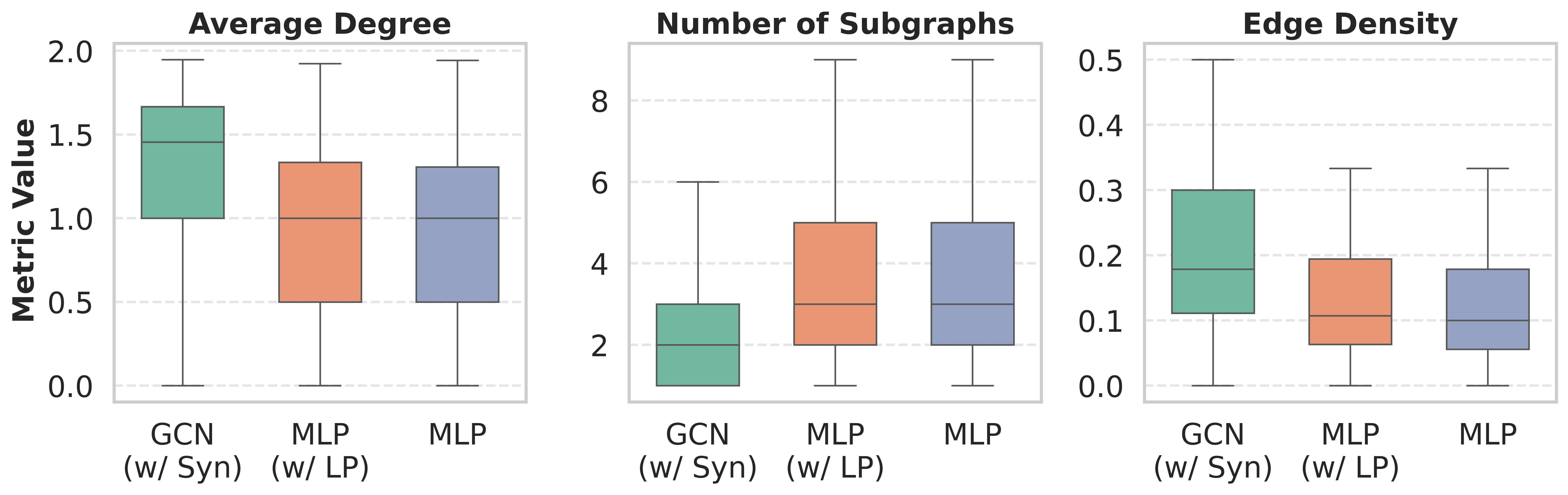}
\caption{\small Comparison of linguistic coherence with respect to (left) average node degree (middle) the number of subgraphs (right) edge density in the HateXplain dataset.}
\label{fig:hatexplain_structure_coherence}
\end{figure*}

\noindent 
\textbf{Linguistic Structure on the \textsc{IMDb} Dataset.}~~
To verify that our observations scale beyond the \textsc{Movies} dataset used in
the main paper, we repeat the structural-coherence ablation on the full
\textsc{IMDb} corpus (25k reviews).  Each selector is constrained to retain
10\% of the input words, and we report three complementary graph-based metrics:

\begin{enumerate}[leftmargin=1.5em, itemsep=0pt]
\item \textbf{Average degree (higher $\uparrow$ is better)}:
In a graph composed of the words (nodes) in a rationale, the number of connected subgraphs measures the degree of token fragmentation. Ideally, this value should be close to one. A single subgraph signifies that all words in the rationale form a single connected component, implying the absence of semantically isolated "loner" tokens. From a structural perspective, this indicates that the model has identified a semantically complete phrase or clause centered around a unified theme or event, rather than simply listing scattered keywords. Therefore, a low number of subgraphs is directly linked to the rationale's high semantic unity.

\item \textbf{Number of connected subgraphs (fewer $\downarrow$ is better)}:
Edge density measures the strength of internal connectivity within a subgraph by comparing the actual number of edges to the maximum possible. High edge density implies that the components of the rationale form a syntactically robust cluster with close, redundant interconnections, rather than a loose chain structure. This indicates that grammatical units with strong dependency relationships in the original sentence, such as subjects and their modifiers or verbs and their objects, have been selected together. A rationale with high edge density exhibits reduced ambiguity as words strongly support each other's meaning, serving as evidence that the model has faithfully preserved the core syntactic relationship network of the sentence, moving beyond superficial connections.

\item \textbf{Edge density (higher $\uparrow$ is better)}:
The average node degree, which represents the average number of edges connected to each node, is a metric for evaluating how well a rationale incorporates the structural backbone of the sentence. In a dependency graph, nodes with a high degree typically act as grammatical "hubs," such as the main verbs or subjects. A high average degree in a rationale suggests that the model has constructed its explanation around these core hubs. This prevents the rationale from consisting solely of peripheral modifiers and ensures the maintenance of the sentence's core semantic framework. In effect, it forces the rationale to be firmly anchored to the core structure of the original sentence, minimizing semantic distortion and ensuring a grammatically stable form.
\end{enumerate}

\noindent\textbf{Results.}  
Figure~\ref{fig:IMDb_structure_coherence} confirms, across
all three structural metrics, that our GCN selector with syntactic edges
(\textbf{GCN~(w/ Syn)}) yields markedly more coherent rationales than either
baseline:

\begin{itemize}[leftmargin=*, topsep=0pt, partopsep=0pt, itemsep=0pt]
    \item \emph{Average degree}.  
          The GCN distribution is centred well above the two baselines
          (median $\approx\!1.2$ compared with $\approx\!0.7$ for \textsc{MLP w/LP} and $\approx\!0.6$ for \textsc{MLP}).  
          Higher per-word connectivity indicates that GCN tends to preserve
          words that are richly interlinked in the knowledge graph.

    \item \emph{Number of subgraphs}.  
          Fewer components are better.  
          GCN rationales typically form a single connected component
          (median $\approx\!7$ subgraphs, upper whisker $\approx\!20$),
          whereas \textsc{MLP w/LP} and \textsc{MLP} fragment the selection
          more severely (medians $\approx\!13$ and $\approx\!14$,
          maxima exceeding 30).  
          The Laplacian penalty reduces fragmentation relative to vanilla MLP but still falls far short of GCN.

    \item \emph{Edge density}.  
          GCN retains the highest proportion of possible edges
          (median $\approx\!0.05$, upper whisker $>\!0.10$),
          almost doubling the density achieved by vanilla MLP and clearly
          surpassing \textsc{MLP w/LP}.  
          Denser subgraphs signal a closer adherence to the underlying
          syntactic relations.
\end{itemize}

\noindent
\textbf{Take-away.}\;
Even with a 15\% selection budget on the large \textbf{IMDb} corpus, only the graph-aware selector maintains high local connectivity, avoids fragmentation, and captures a greater share of syntactic edges. 
Post-hoc Laplacian regularisation helps the MLP baseline, but cannot match the
structural faithfulness achieved by explicitly modelling linguistic graphs during selection. Results for the HateXplain dataset are in Figure~\ref{fig:hatexplain_structure_coherence}, respectively.

\noindent
\textbf{Note on \textsc{Graph-SST2}.}  
We omit \textsc{Graph-SST2} from this analysis because its sentences are
typically single clauses (avg.\ 10.2 nodes), leading every selector to
produce a fully connected subgraph; the resulting distributions degenerate and
offer little comparative insight.

\noindent
These additional results reinforce the main paper conclusion: leveraging linguistic graph structure \emph{during} selection --- Rather than enforcing connectivity through post-hoc methods --- is crucial for extracting rationales that faithfully preserve the syntactic relationships of the original text.

\section{Additional Experiments} \label{appendix: add_expriments}

\subsection{API-Based Model Evaluation: Keyword Extraction} \label{appendix:api_expriments}

To validate GLASS's applicability in real-world API scenarios where only input-output access is available, we conduct experiments on the Inspec dataset for keyword extraction using a commercial API-based language model.

\noindent 
\textbf{Dataset and Setup.}~~
The Inspec~\cite{hulth_inspec_EMNLP2003} dataset contains scientific abstracts from the computer science and information technology domains, each accompanied by human-annotated keywords that capture the core topics and concepts. The keywords are typically technical terms and domain-specific phrases that may not appear verbatim in the abstract, requiring models to identify semantically central content rather than simply extracting frequent words. This makes it an ideal benchmark for evaluating explanation methods in specialized domains where identifying salient technical concepts is crucial.

We use a Hugging Face API-based model, DistilBERT-Inspec \cite{kulkarni_inspecBERT_2022}, where we only have access to input text queries and output predictions --- i.e., no gradients, embeddings, logits, or internal model states are accessible. This simulates typical production API deployment scenarios where users interact solely through input-output pairs. For the embedding model ($f$), we extract representations from the penultimate layer, where the reward is defined as the F1 score measuring overlap between keywords extracted from the selected word subset and ground-truth annotations. This setup replicates typical production API scenarios where only input-output access is available.

\noindent 
\textbf{Results.}~~
Table~\ref{tab:inspec} shows that GLASS achieves an F1 score of 0.441 ± 0.004 only using 30\% of the input text, which is 90.2\% of the oracle performance with full text (F1 score: 0.489). Here, the F1 score measures the overlap between the extracted keywords and the ground-truth annotations. This demonstrates that: (i) GLASS is fully compatible with strict API-based black-box settings, (ii) our method can effectively identify informative word subsets even without model internals, and (iii) the approach generalizes to diverse generation tasks beyond summarization.

\subsection{Effect of Selection Rate on Summarization Performance} \label{appendix: summarization_experiment}
\begin{figure}[h!]
\centering
\includegraphics[trim={0cm 0cm 0cm 0cm}, clip, width=0.99\linewidth]{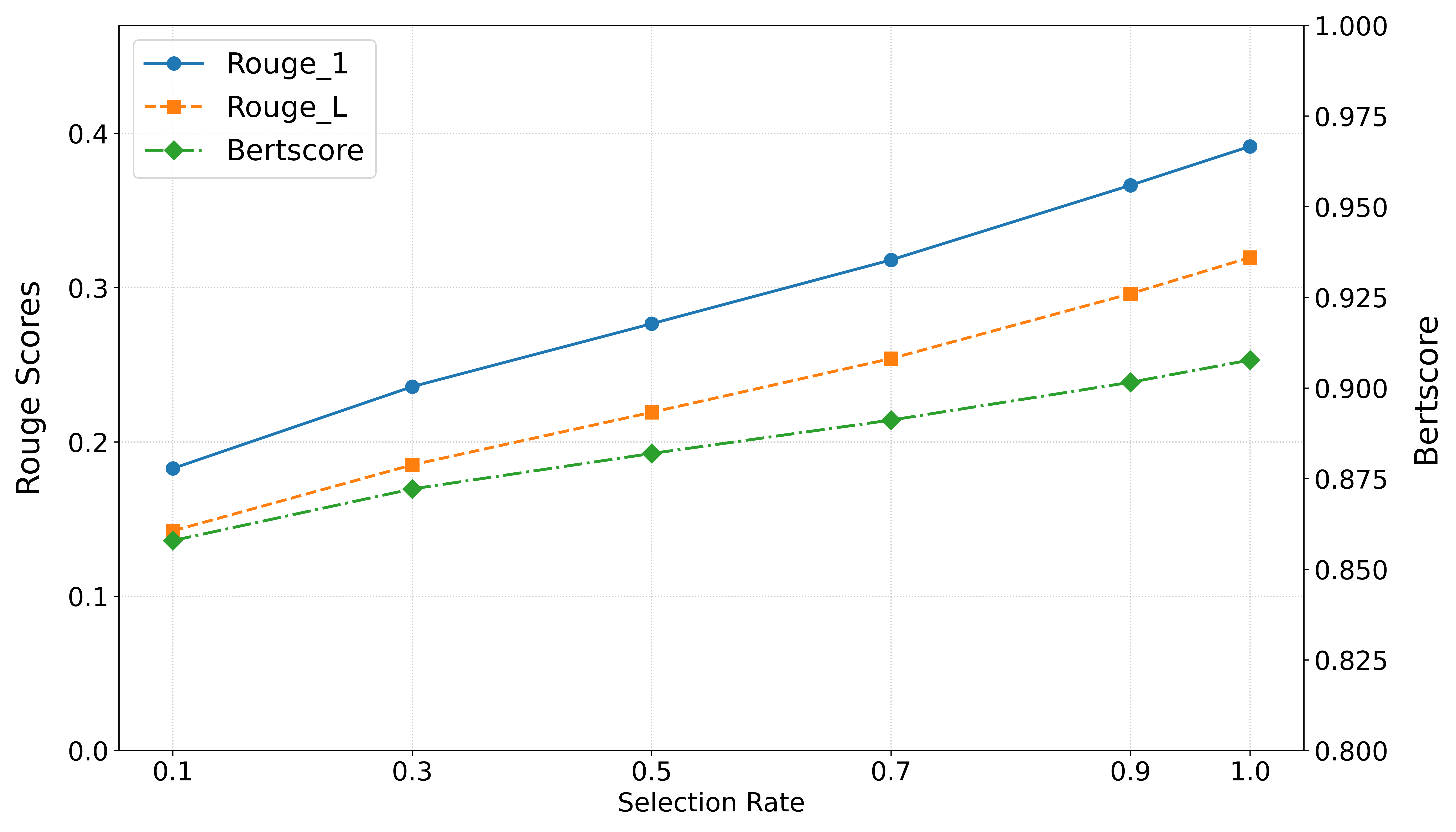}
\caption{\small Performance Changes of Rouge-1, Rouge-L, and Bertscore Metrics According to Selection Rate. The left Y-axis represents Rouge scores, while the right Y-axis represents Bertscore, visualizing their different value ranges. All metrics generally show a linear decrease in performance as the Selection Rate decreases.}
\Description{XSum selection rate variation}
\label{fig:xsum_figure}
\end{figure}

To investigate how selection rate impacts performance in a generative setting, we analyze results on the \textbf{XSUM} dataset under varying selection rates. As shown in Figure~\ref{fig:xsum_figure}, all metrics—Rouge-1, Rouge-L, and BertScore—consistently decline as the selection rate decreases. This trend is expected, as reducing the number of available input tokens limits the information accessible to the summarization model. Notably, the performance drop appears roughly linear across the range of selection rates, suggesting that our model degrades gracefully even under strong sparsity constraints. The dual-axis visualization emphasizes the difference in absolute value ranges between the Rouge and BertScore metrics, while reinforcing the same downward trend across evaluation criteria.

\noindent 
\textbf{Additional Qualitative Analysis -- Explanations Based on Word Subsets.}~~
We compare explanations produced by our GCN selector (\textbf{GLASS}) and two
baselines --- an MLP and an MLP with Laplacian regularization
(\textsc{MLP}\, w/\textsc{LP}) --- on two representative reviews from the Movies dataset, which contains human‐annotated rationales. For fairness, every selector is constrained to keep 15 \% of the input words.

\subparagraph{Sample 1 \,--\, Figure~\ref{fig:Word_subsets_explanation_15}\,(a–c)}

\begin{itemize}[leftmargin=1.2em]

\item \textbf{\textit{“this is a marvelous confection — cocky, funny, thought provoking.”}}

      GCN keeps the contiguous span  
      \emph{“is a marvelous confection — cocky, funny, thought provoking”},  
      reflecting the coordination of three adjectives and preserving the dash that signals an explanatory aside.  
      By contrast, \textsc{MLP} and \textsc{MLP}\, w/\textsc{LP} shrink the clause to the fragment \emph{“is a marvelous confection thought.”},
      deleting both punctuation and two adjectives.  
      The resulting phrase is ungrammatical and dilutes the strongly positive sentiment of the original sentence.

\item \textbf{\textit{“the film is something else, something original.”}}

      GCN again retains the full coordinated structure  
      \emph{“is something else, something original”}.  
      The two MLP baselines drop the first \emph{something} and the comma,
      yielding \emph{“, is something else original”}, a fragment that breaks the
      intended parallelism and can even be misread as negative when taken out of context.  This illustrates how losing even a single connective can distort the rhetorical force of a sentence.
\end{itemize}

\subparagraph{Sample 2 \,--\, Figure~\ref{fig:Word_subsets_explanation_15}\,(d–f)}
\begin{itemize}[leftmargin=*, topsep=0pt, partopsep=0pt, itemsep=0pt]
\item \textbf{\textit{“The dialogue is sharp and well written, with many funny set pieces.”}}

      GCN selects  
      \emph{“is sharp and well written, with many funny set pieces”},  
      a fluent clause that includes the conjunction \emph{and} and the adjunct
      prepositional phrase.  
      The MLP baseline chooses \emph{“dialogue is sharp well, funny,”} while
      \textsc{MLP}\, w/\textsc{LP} yields \emph{“is sharp well, funny.”}
      Both lose the coordination and misplace \emph{well}, generating a
      disfluent fragment that weakens the positive tone.

\item \textbf{\textit{“Sevigny delivers an honest performance, and makes the audience feel for her character.”}}

      GCN captures the graph-connected span  
      \emph{“delivers an honest performance makes audience feel,”} preserving
      both coordinated predicates and the sentiment-bearing object
      \emph{audience}.  
      The MLP baseline keeps only the first clause
      \emph{“delivers an honest performance,”} while \textsc{MLP}\, w/\textsc{LP}
      retains the disfluent string \emph{“delivers an honest performance makes
      for,”} omitting the key verb \emph{feel}.  Both fail to express the causal
      link that underpins the review’s positive judgment.
\end{itemize}

\noindent
Overall, Samples 1, 2 shows that post-hoc grammatical regularization alone cannot stop the MLP variants from producing fragmented, sentiment-diluting rationales, whereas explicitly modeling graph edges enables \textbf{GLASS} to keep well-formed, label-aligned spans.

\subparagraph{Sample 3 -- 20 \% Selection (Figs.~\ref{fig:Word_subsets_explanation_20_part1})}
\begin{itemize}[leftmargin=*, topsep=0pt, partopsep=0pt, itemsep=0pt]

\item \textbf{\textit{“I believe that Robert Duvall (who is the producer, director, writer, and main star of  The Apostle) deserves an Oscar for his performance as Sonny the religious crusader — a performance which is so complex and realistic it ranks as one of the finest acting performances on film.”}}

      GCN retains the core predicate chain
      (\emph{believe … Duvall … deserves … performance … is so complex and
      realistic it ranks as one of finest}), keeping both evaluative verbs
      \emph{deserves} and \emph{ranks} as well as the sentiment-bearing adjectives \emph{complex} and \emph{realistic}.  Although certain function words are omitted, the resulting span is fluent and clearly conveys the reviewer’s high praise.

      \textsc{MLP} drops the head verb \emph{deserves}, scrambles word order,
      and splits modifiers from their host noun, yielding the disfluent fragment
      \emph{“believe that Duvall for as performance is complex realistic ranks
      as one of finest.”}  The intended sentiment becomes opaque.

      \textsc{MLP}\, w/\textsc{LP} restores the intensifier phrase
      \emph{“so complex and realistic”}, but still omits \emph{deserves},
      misplaces \emph{for}, and leaves the span without a grammatical subject,
      producing \emph{“believe Duvall for is so complex and realistic ranks as
      one of the finest.”}  Despite the Laplacian penalty, key syntactic links
      remain broken.

\end{itemize}

\noindent
\textbf{Observation.}\;
Increasing the word budget from 15 \% to 20 \% --- a one-third expansion of the
allowance --- does not rescue the two MLP baselines: they still yield fragmented,
partially misplaced words that blur the review’s positive stance.  By contrast, our graph-aware selector (\textbf{GLASS}) continues to produce a contiguous, sentiment-aligned span that preserves the long-range dependencies among \emph{deserves}, \emph{performance}, and \emph{ranks}, confirming the robustness of incorporating linguistic-graph structure into the selection process.

\noindent 
\textbf{Additional Qualitative Analysis -- HateXplain Case Study.}~~
We further analyze the explanations produced by our GCN-based selector (\textbf{GLASS}) and the MLP baseline on two representative examples from the HateXplain dataset. For a fair comparison, both models are constrained to retain 30\% of input tokens. Despite comparable annotation precision scores (0.750 vs. 0.800), we observe key differences in linguistic coherence and structural alignment.

\subparagraph{Sample 1 \,--\, Figure~\ref{fig:Word_subsets_explanation_hatexplain}\,(a–c)}
\begin{itemize}[leftmargin=1.2em]

\item \textbf{\textit{“broke jews control the world woke ... russia is funding the alt right and jared taylor ... movement boomerlivesdontmatter”}}

      The GCN-based selector (\textbf{GLASS}) selects semantically connected phrases such as  
      \emph{“funding the movement”} and \emph{“boomerlivesdontmatter”}, preserving the structural relations that underpin the causal logic of the hate speech claim. Despite a slightly lower annotation precision (0.750), the selected tokens form a syntactically valid subgraph aligned with the underlying dependency structure. 

      In contrast, the MLP selector shows fragmented selections, such as  
      \emph{“funding,” “movement,” “foot,” “generation,” “control”}, that are disconnected in the graph and fail to maintain grammatical relationships. The resulting explanation lacks cohesion and may distort the intended rhetorical flow, even though it achieves slightly higher annotation precision (0.800).

\end{itemize}

\subparagraph{Sample 2 \,--\, Figure~\ref{fig:Word_subsets_explanation_hatexplain}\,(d-f)}
\begin{itemize}[leftmargin=1.2em]

\item \textbf{\textit{“the catholic church stopped being a bulwark against western rot when the church became infested with homosexuals …”}}

      \textbf{GLASS} demonstrates perfect alignment with the human annotation (precision 1.000), capturing the coherent clause  
      \emph{“church became infested with homosexuals”} as a complete semantic unit. It also selects other structurally important spans such as  
      \emph{“bulwark against western rot”} and  
      \emph{“depraved gay orgies on sacred church ground”}, which include sentiment-bearing terms and syntactic dependencies. These selections form tightly connected subgraphs in the dependency structure, preserving the original argument’s logic and emotional tone.

      Meanwhile, the MLP baseline fragments this structure, dropping the subject \emph{“church”} and selecting disjointed tokens like \emph{“corrupted,” “bureaucracies,” “stories”}, breaking the causal connection that drives the review’s central claim. The explanation becomes semantically ambiguous and less informative.

\end{itemize}

\noindent
\textbf{Observation.}\;
Even under identical selection budgets, our graph-aware selector produces linguistically fluent and semantically coherent rationales that preserve key argumentative structures. In contrast, MLP-based methods generate fragmented outputs that dilute interpretability and disrupt grammatical flow. These examples underscore the advantage of integrating syntactic structure for faithful and human-aligned explanation generation.

\noindent 
\textbf{Additional Qualitative Analysis -- XSum Case Study.}~~
We further analyze the important word subsets produced by our graph-based selector (\textbf{Graph-Selector}). The analysis focuses on how the selector preserves linguistic coherence and logical structure, moving beyond simple keyword extraction to capture the underlying meaning of the text.

\begin{figure*}[ht]
  \begin{subfigure}[b]{0.825\textwidth}
    \includegraphics[trim={0cm 0cm 0cm 0cm}, clip, width=\linewidth]{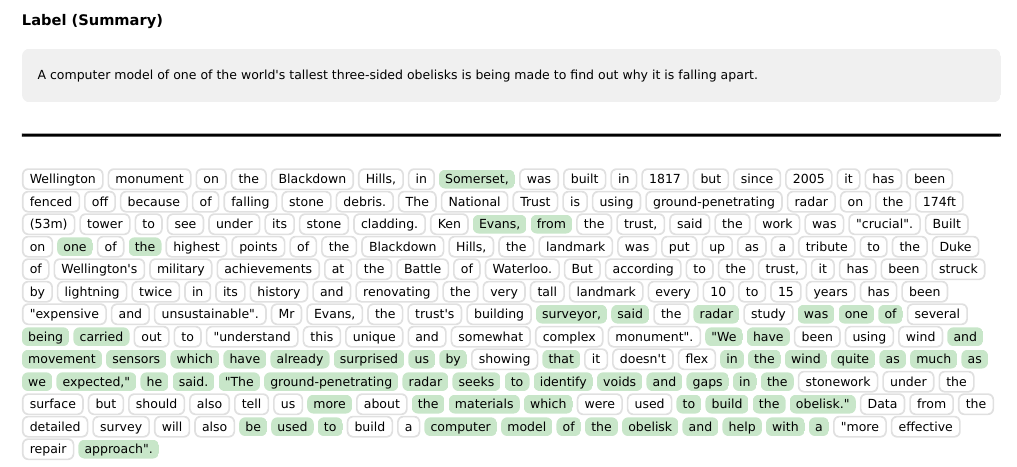}
  \end{subfigure}
  \begin{subfigure}[b]{0.8\textwidth}
    \includegraphics[trim={0cm 0cm 0cm 0cm}, clip, width=\linewidth]{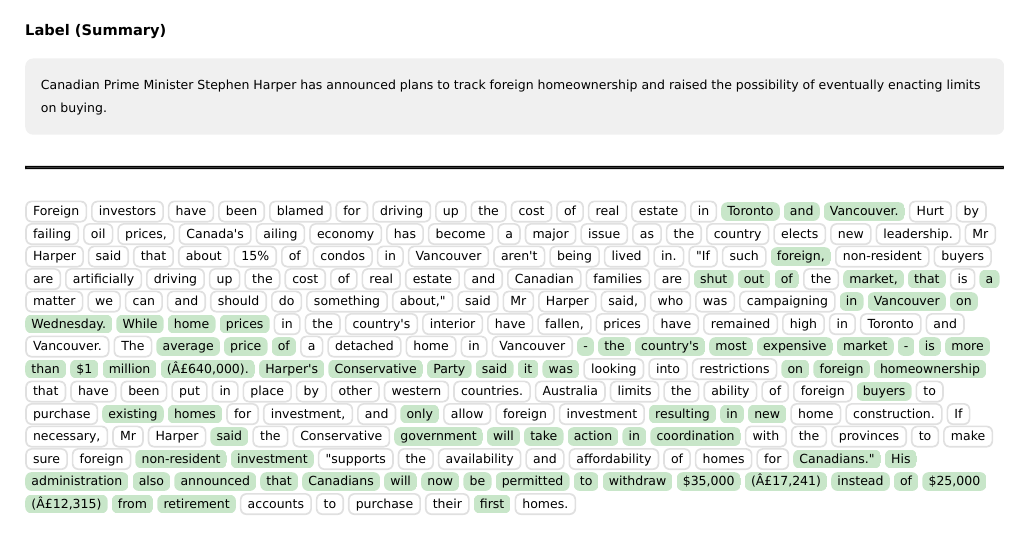}
  \end{subfigure}
    \caption{Examples of important word subsets identified by our proposed method on the XSum dataset. Words highlighted in \textbf{green} are those selected by our model for summary generation.}
  \label{fig:Word_subsets_explanation_xsum}
  \Description{Wellington Monument, Canadian Real Estate}
\end{figure*}

\subparagraph{Sample 1\,--\, Figure~\ref{fig:Word_subsets_explanation_xsum}}
\begin{itemize}[leftmargin=1.2em]

\item \textbf{\textit{“...Wellington monument ... has been fenced off because of falling stone debris... using ground-penetrating radar... to build a computer model of the obelisk and help with a 'more effective repair approach'.”}}

    The \proposed selects semantically connected phrases such as 
    \emph{“voids and gaps in the stonework”} and \emph{“build a computer model of the obelisk.”} This approach preserves the structural relations that underpin the procedural logic of the article (Problem $\rightarrow$ Method $\rightarrow$ Goal). The selected tokens form a syntactically valid subgraph that captures the core technical narrative.

\end{itemize}

\subparagraph{Sample 2\,--\, Figure~\ref{fig:Word_subsets_explanation_xsum}}
\begin{itemize}[leftmargin=1.2em]

\item \textbf{\textit{“...foreign investors have been blamed for driving up the cost of real estate... Harper's Conservative Party said it was looking into restrictions on foreign homeownership...”}}

    The \proposed demonstrates a strong ability to identify and preserve complete semantic units. It captures the coherent clause 
    \emph{“Harper's Conservative Party said it was looking into restrictions on foreign homeownership,”} which forms the backbone of the final summary. It also selects other structurally important spans that provide context, such as 
    \emph{“shut out of the market,”} preserving the original argument’s logic regarding cause and effect.

\end{itemize}

\begin{figure*}[ht]
  \centering
  \begin{subfigure}[b]{0.8\textwidth}
    \includegraphics[trim={0cm 0cm 0cm 0cm}, clip, width=\linewidth]{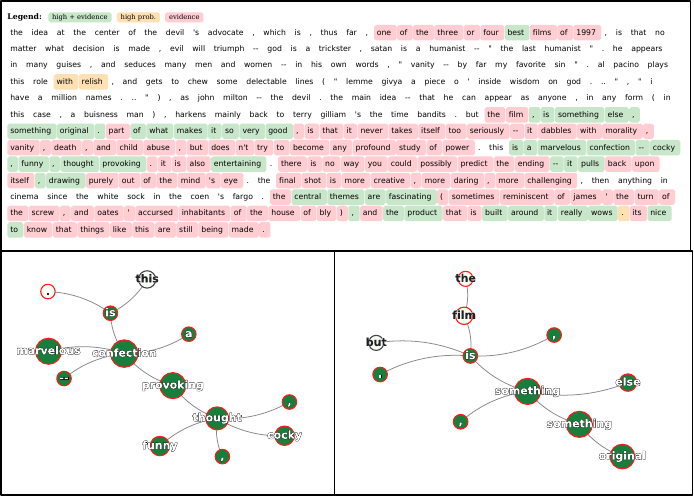}
    \caption{\small GCN w/ syntactic dependency graph-15\% (Annotation Precision: 0.940)}
  \end{subfigure}
    \hfill
  \begin{subfigure}[b]{0.8\textwidth}
    \includegraphics[trim={0cm 0cm 0cm 0cm}, clip, width=\linewidth]{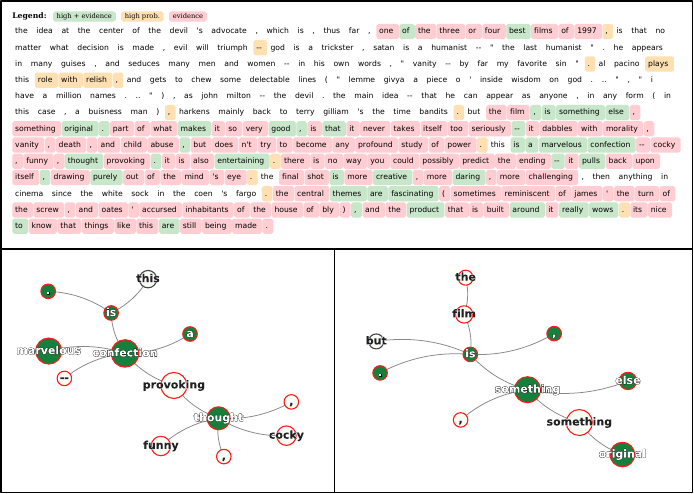}
    \caption{\small MLP-15\% (Annotation Precision: 0.716)}
  \end{subfigure}
  \Description{mlp example}
  \end{figure*}
\begin{figure*}[ht]
\ContinuedFloat
  \begin{subfigure}[b]{0.8\textwidth}
    \includegraphics[trim={0cm 0cm 0cm 0cm}, clip, width=\linewidth]{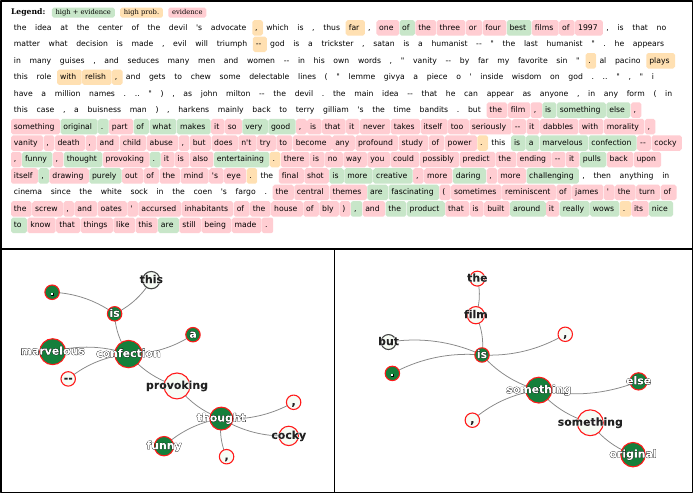}
    \caption{\small MLP w/ graph Laplacian-15\% (Annotation Precision: 0.760)}
  \end{subfigure}
\ContinuedFloat
\begin{subfigure}[b]{0.8\textwidth}
    \includegraphics[trim={0cm 0cm 0cm 0cm}, clip, width=\linewidth]{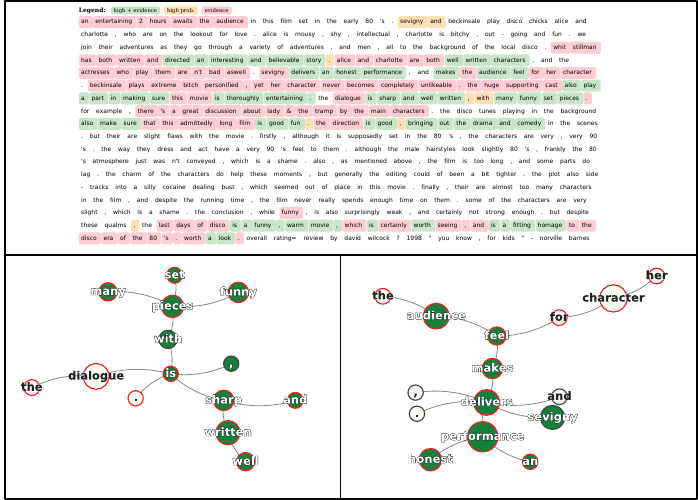}
    \caption{\small GCN w/ syntactic dependency graph-15\% (Annotation Precision: 0.890)}
  \end{subfigure}
  \Description{GCN example}
\end{figure*}
\begin{figure*}[ht]
\ContinuedFloat
  \begin{subfigure}[b]{0.8\textwidth}
    \includegraphics[trim={0cm 0cm 0cm 0cm}, clip, width=\linewidth]{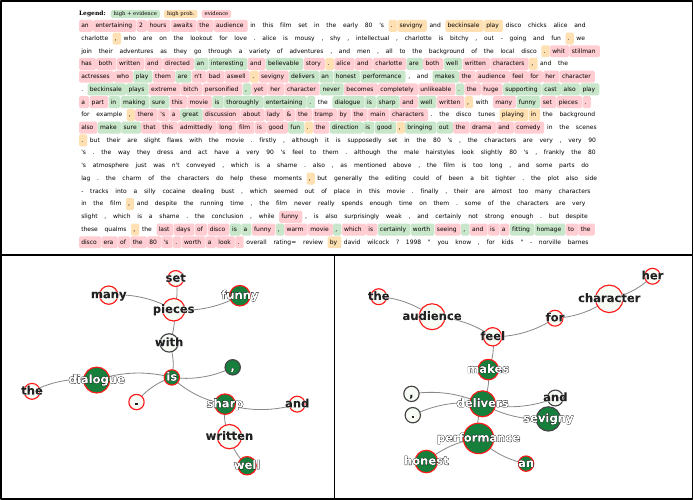}
    \caption{\small MLP-15\% (Annotation Precision: 0.708)}
  \end{subfigure}
  \hfill
  \begin{subfigure}[b]{0.8\textwidth}
    \includegraphics[trim={0cm 0cm 0cm 0cm}, clip, width=\linewidth]{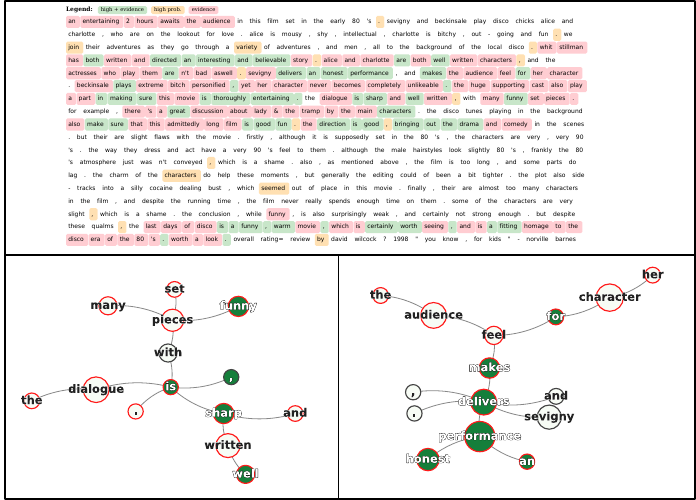}
    \caption{\small MLP w/ graph Laplacian-15\% (Annotation Precision: 0.767)}
  \end{subfigure}
  \caption{\small Examples of important word subsets identified by our proposed method on the Movies dataset. Based on the human rationales provided in the dataset, words highlighted in \hrationale{green} correspond to selected human rationales, those in \hmissed{red} represent unselected human rationales, and words highlighted in \hextra{orange} denote selected words that were not part of the human rationale. The accompanying subgraph illustrates the structural relationships among selected words, highlighting how well the explanation aligns with the underlying linguistic structure. We also report precision and recall scores relative to the human rationales to quantify alignment.}
  \label{fig:Word_subsets_explanation_15}
  \Description{Movies dataset Word_subsets_explanation}
\end{figure*}

\begin{figure*}[!htbp]
  \centering
  \begin{subfigure}[b]{0.8\textwidth}
    \includegraphics[trim={0cm 0cm 0cm 0cm}, clip, width=\linewidth]{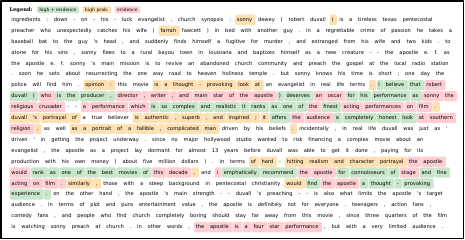}
    \includegraphics[trim={0cm 0cm 0cm 0cm}, clip, width=\linewidth]{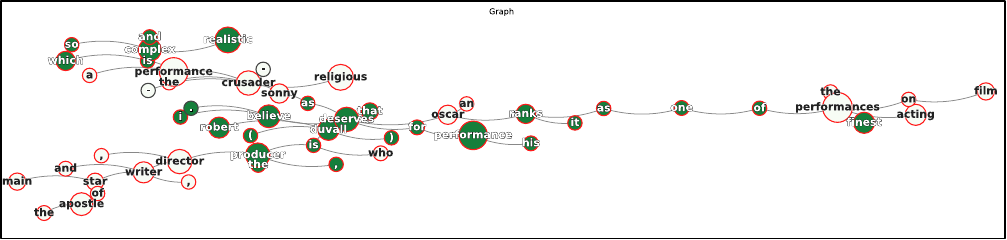}
    \caption{\small GCN w/ syntactic dependency graph-20\% (Annotation Precision: 0.530)}
  \end{subfigure}
  \begin{subfigure}[b]{0.8\textwidth}
    \includegraphics[trim={0cm 0cm 0cm 0cm}, clip, width=\linewidth]{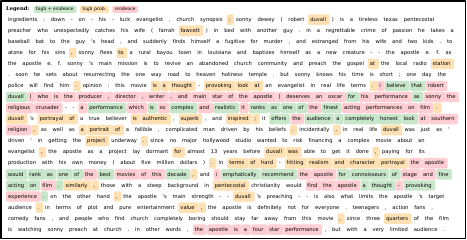}
    \includegraphics[trim={0cm 0cm 0cm 0cm}, clip, width=\linewidth]{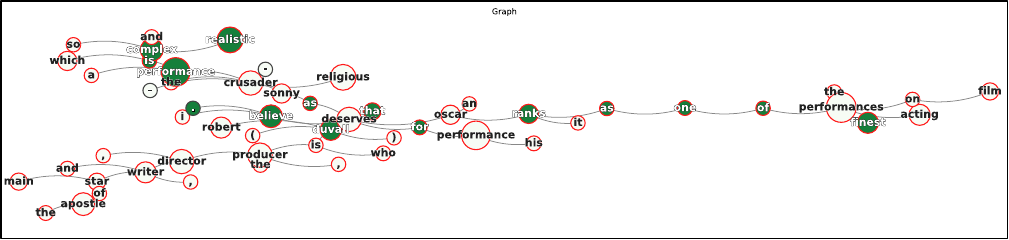}
    \caption{\small MLP-20\% (Annotation Precision: 0.408)}
  \end{subfigure}
  \caption{Additional experiments in 20\% selections (Part 1)}
  \label{fig:Word_subsets_explanation_20_part1}
  \Description{Word_subsets_explanation_20_part1}
\end{figure*}
\begin{figure*}[!htbp]
  \ContinuedFloat
  \centering
  \begin{subfigure}[b]{0.76\textwidth}
    \includegraphics[trim={0cm 0cm 0cm 0cm}, clip, width=\linewidth]{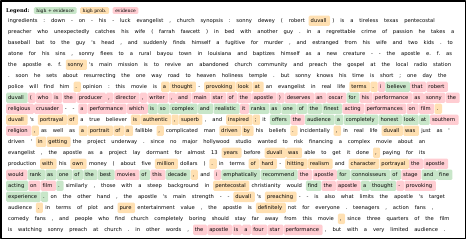}
    \includegraphics[trim={0cm 0cm 0cm 0cm}, clip, width=\linewidth]{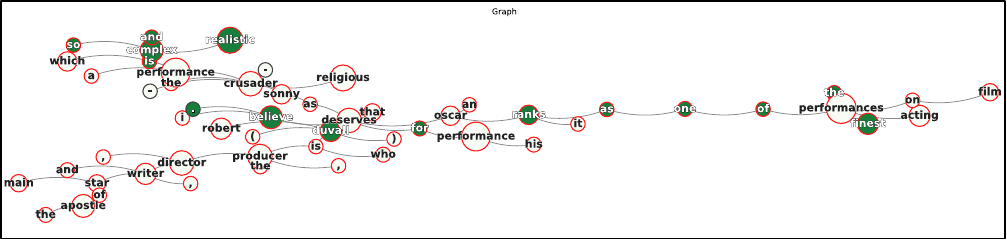}
    \caption{\small MLP w/ graph Laplacian-20\% (Annotation Precision: 0.434)}
  \end{subfigure}
  \caption{Additional experiments in 20\% selections (Part 2)}
  \label{fig:Word_subsets_explanation_20_part2}
  \Description{Word_subsets_explanation_20_part2}
\end{figure*}

\begin{figure*}[ht]
  \centering
  \begin{subfigure}[b]{0.7\textwidth}
    \includegraphics[trim={0cm 0cm 0cm 0cm}, clip, width=\linewidth]{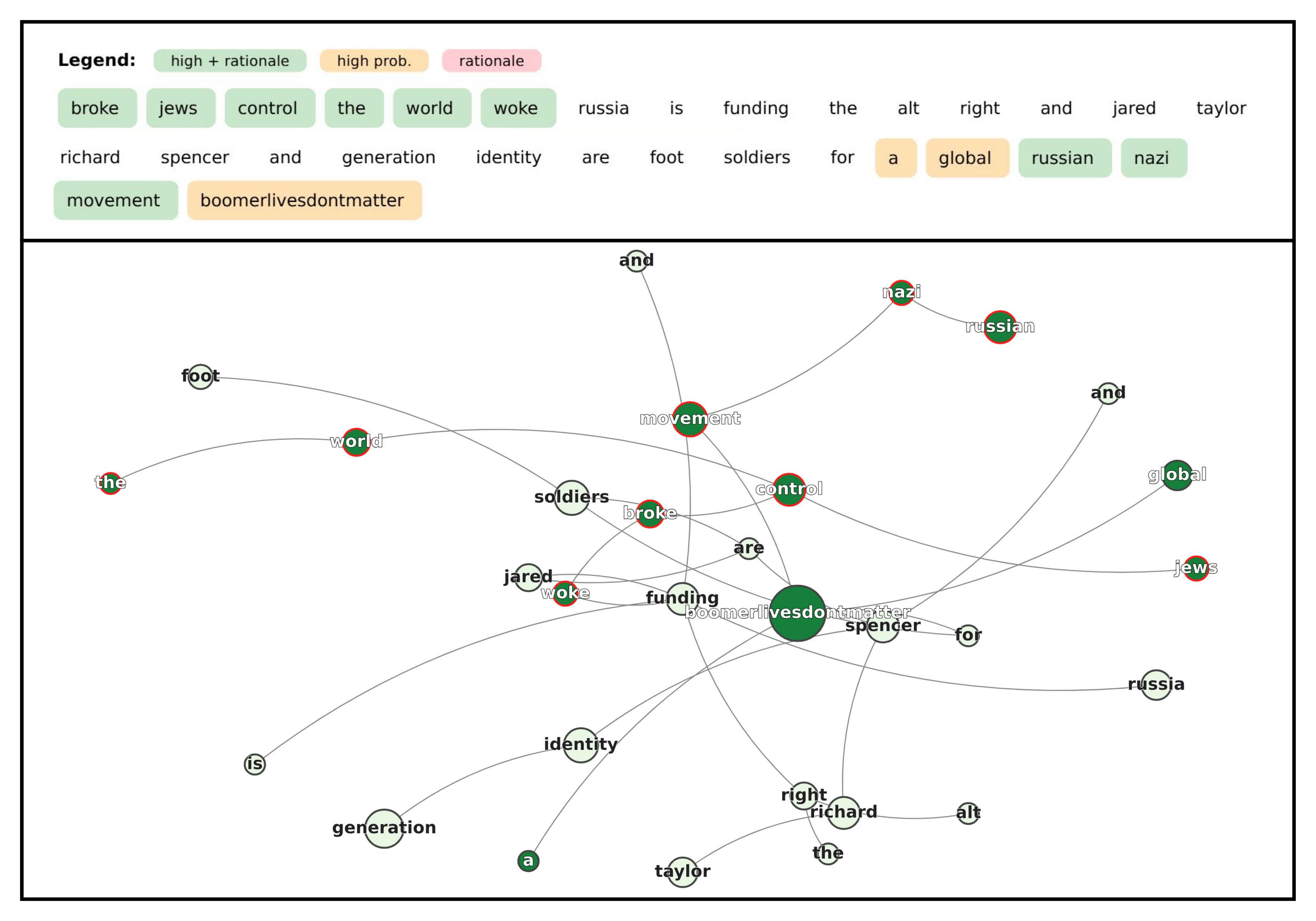}
    \caption{\small GCN w/ syntactic dependency graph-30\% (Annotation Precision: 0.750)}
  \end{subfigure}
  \Description{GCN example}
\end{figure*}
\begin{figure*}[ht]
\ContinuedFloat
  \begin{subfigure}[b]{0.75\textwidth}
    \includegraphics[trim={0cm 0cm 0cm 0cm}, clip, width=\linewidth]{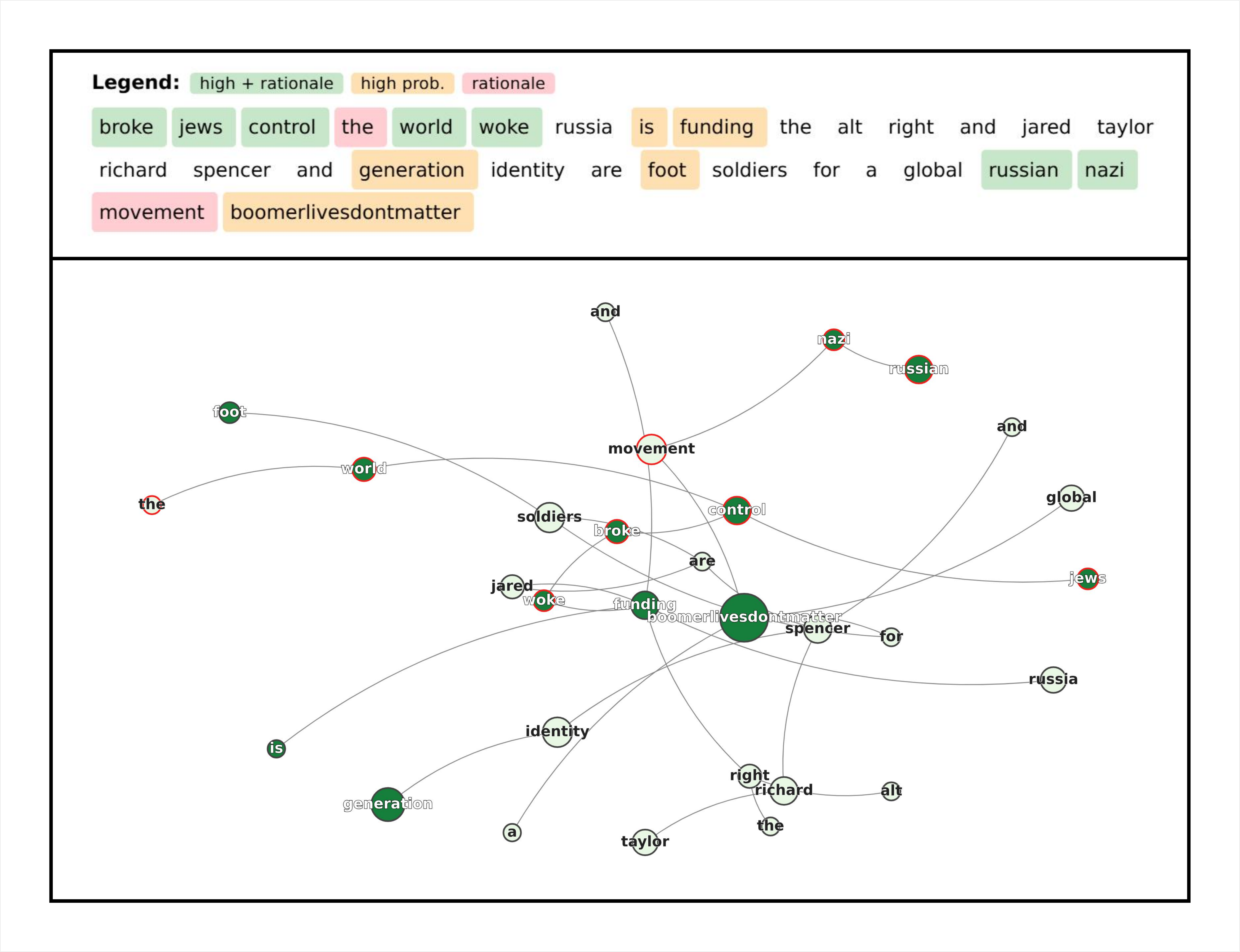}
    \caption{\small MLP-30\% (Annotation Precision: 0.80)}
  \end{subfigure}

  \begin{subfigure}[b]{0.75\textwidth}
    \includegraphics[trim={0cm 0cm 0cm 0cm}, clip, width=\linewidth]{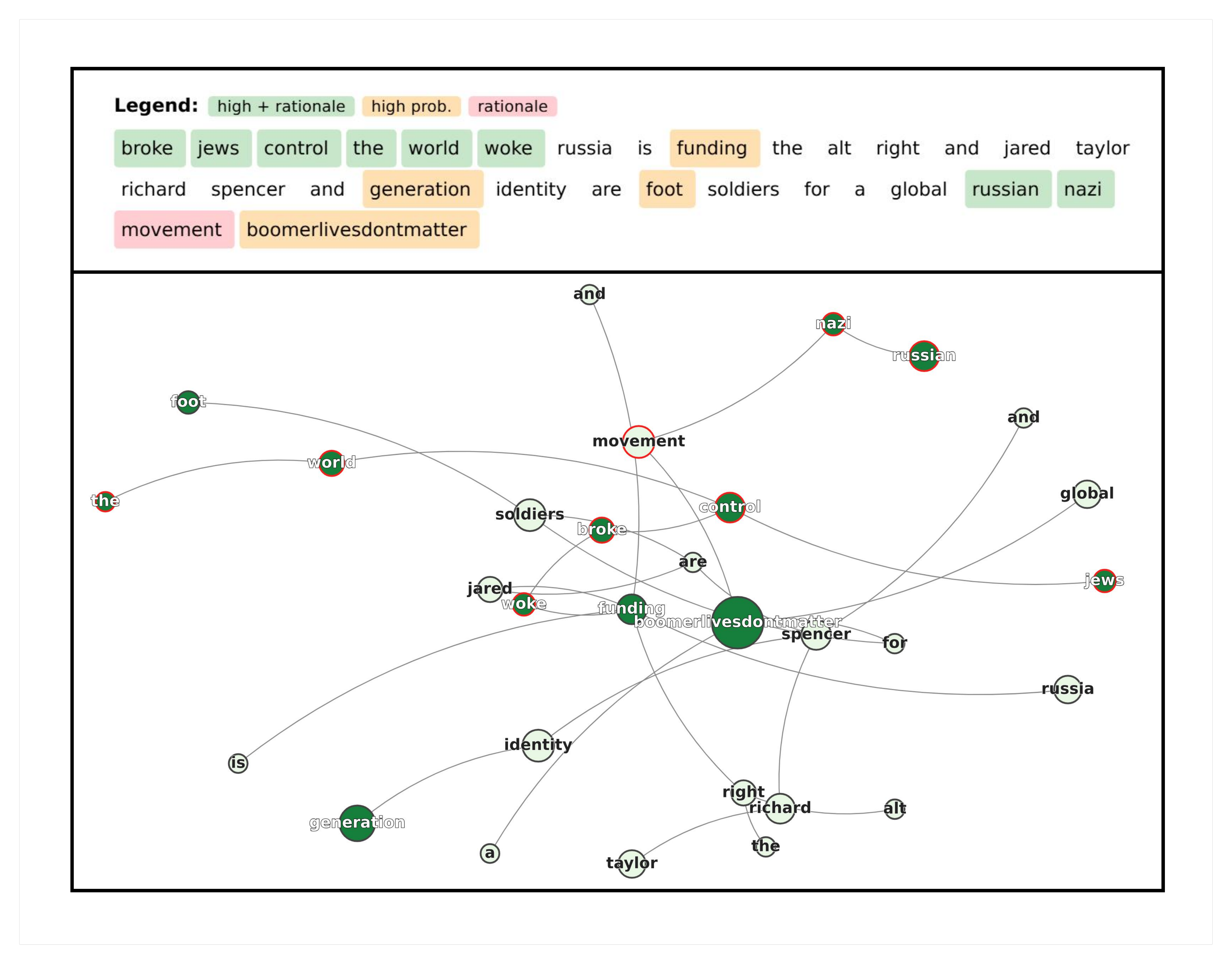}
    \caption{\small MLP w/ graph Laplacian-30\% (Annotation Precision: 0.600)}
    \Description{MLP w/ graph Laplacian example}
  \end{subfigure}
\end{figure*}
\begin{figure*}[ht]
\ContinuedFloat
  \begin{subfigure}[b]{0.85\textwidth}
    \includegraphics[trim={0cm 0cm 0cm 0cm}, clip, width=\linewidth]{figures/hatexplain_GCN_27.pdf}
    \caption{\small GCN-30\% (Annotation Precision: 1.000)}
  \end{subfigure}

  \begin{subfigure}[b]{0.83\textwidth}
    \includegraphics[trim={0cm 0cm 0cm 0cm}, clip, width=\linewidth]{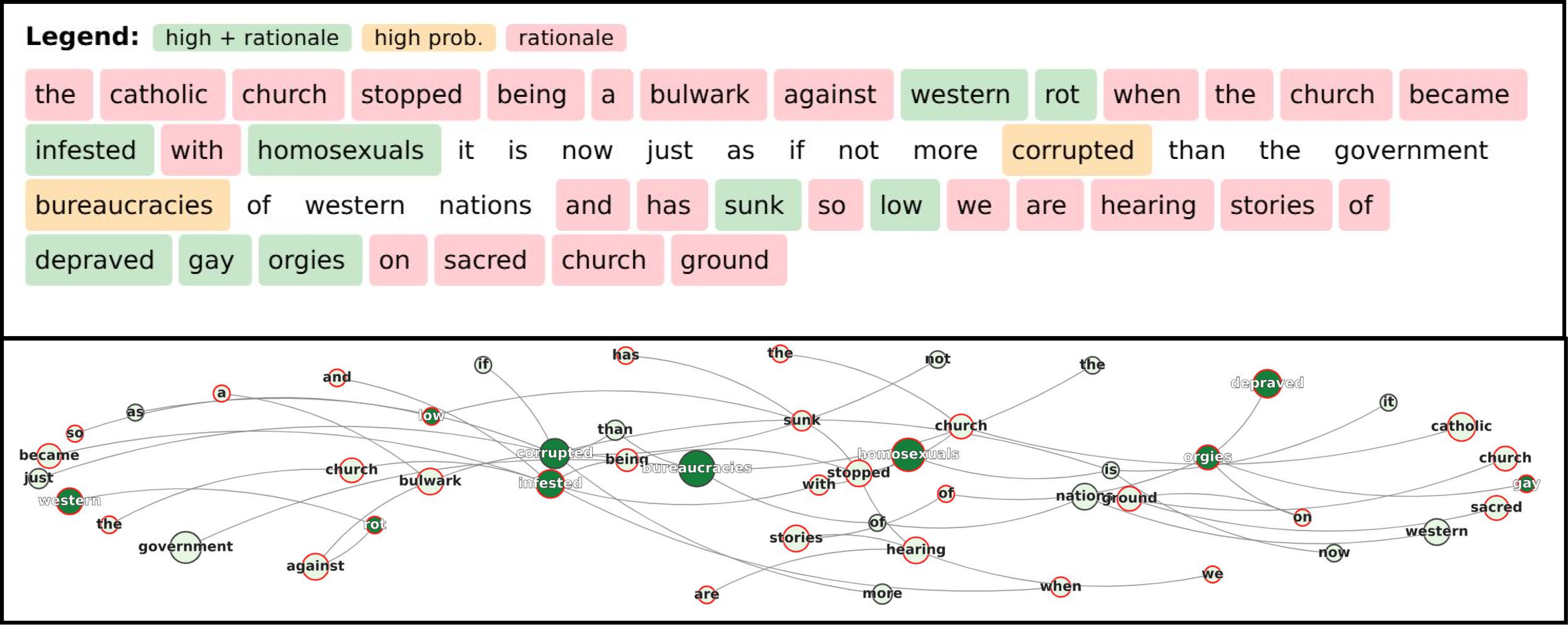}
    \caption{\small MLP-30\% (Annotation Precision: 0.830)}
  \end{subfigure}

  \begin{subfigure}[b]{0.83\textwidth}
    \includegraphics[trim={0cm 0cm 0cm 0cm}, clip, width=\linewidth]{figures/hatexplain_MLP_LP_27.pdf}
    \caption{\small MLP w/ graph Laplacian-30\% (Annotation Precision: 0.667)}
  \end{subfigure}
  \caption{Examples of important word subsets identified by our proposed method on the HateXplain dataset}
  \label{fig:Word_subsets_explanation_hatexplain}
  \Description{Word_subsets_explanation_hatexplain}
\end{figure*}

\begin{figure*}[!t]
  \centering
  \begin{subfigure}{\textwidth}
    \centering
    \includegraphics[width=1.0\linewidth]{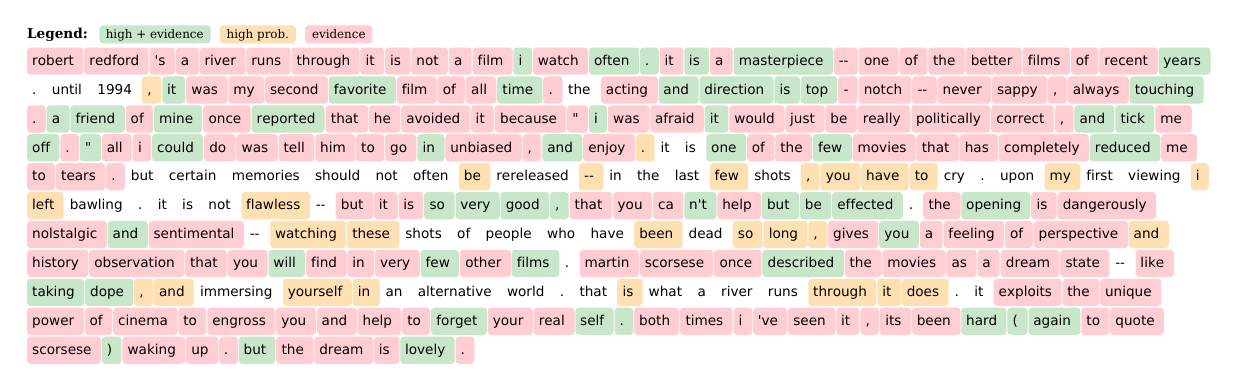}
    \caption*{\small SHAP (Annotation Precision: 0.671)}
  \end{subfigure}

  \begin{subfigure}{\textwidth}
    \centering
    \includegraphics[width=1.0\linewidth]{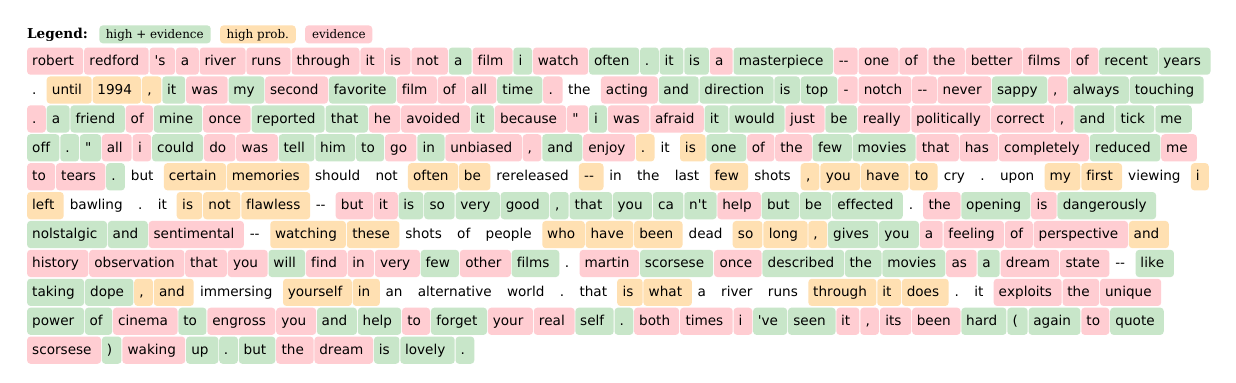}
    \caption*{\small LIME (Annotation Precision: 0.694)}
  \end{subfigure}

  \begin{subfigure}{\textwidth}
    \centering
    \includegraphics[width=1.0\linewidth]{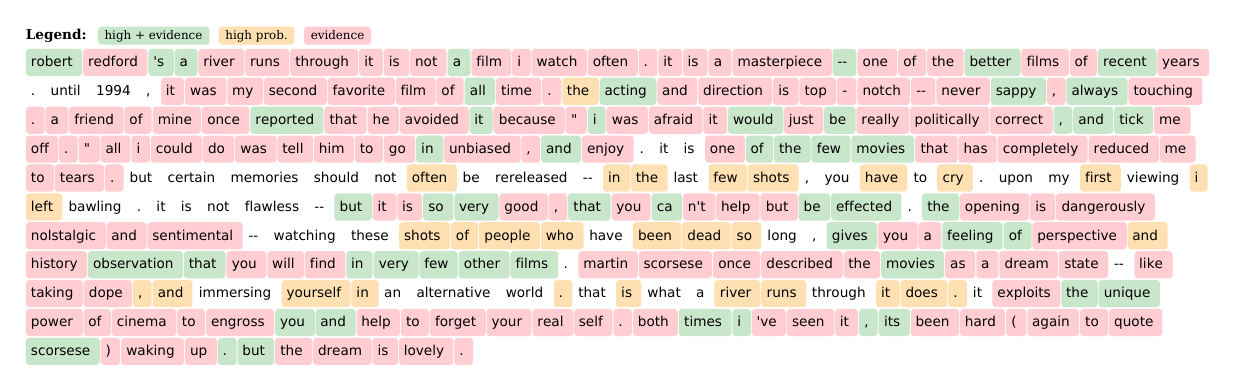}
    \caption*{\small IG (Annotation Precision: 0.649)}
  \end{subfigure}

\end{figure*}

\begin{figure*}[!t]

  \begin{subfigure}{\textwidth}
    \centering
    \includegraphics[width=1.0\linewidth]{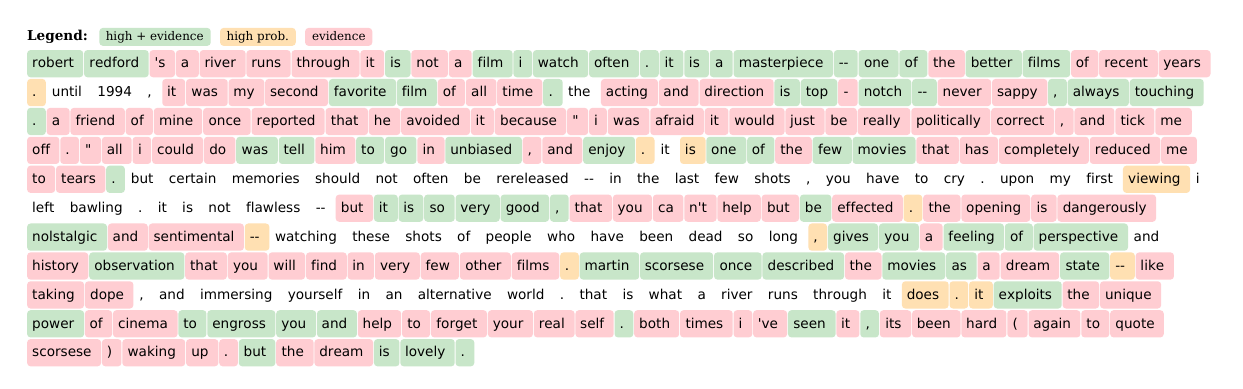}
    \caption*{\small L2X (Annotation Precision: 0.718)}
  \end{subfigure}

  \begin{subfigure}{\textwidth}
    \centering
    \includegraphics[width=1.0\linewidth]{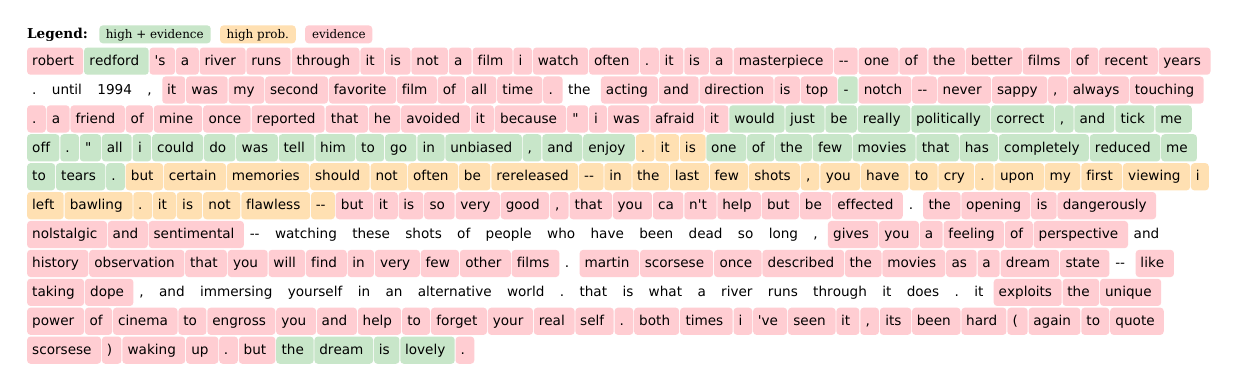}
    \caption*{\small LTX (Annotation Precision: 0.590)}
  \end{subfigure}

  \begin{subfigure}{\textwidth}
    \centering
    \includegraphics[width=1.0\linewidth]{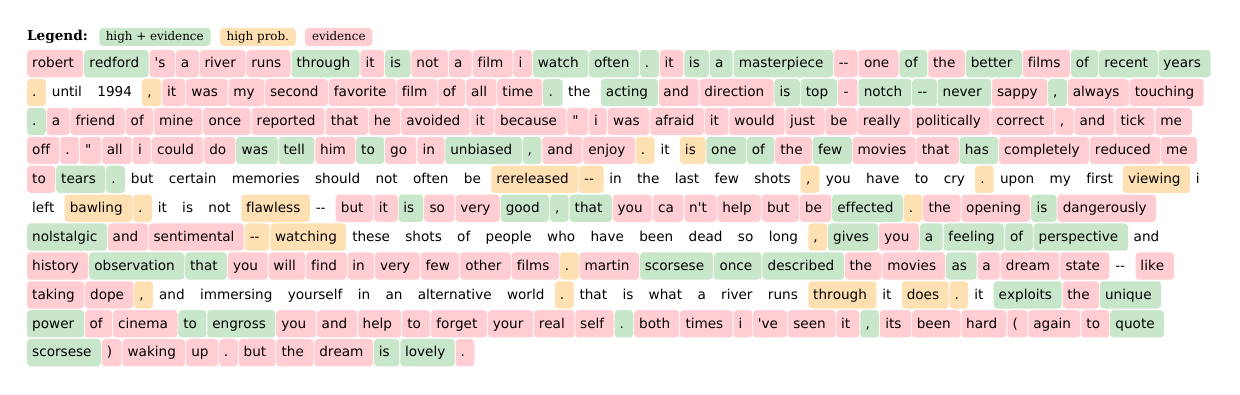}
    \caption*{\small CXPlain (Annotation Precision: 0.725)}
  \end{subfigure}

  \caption{\small Baseline explanation methods on the same sample as 
Figure~\ref{fig:movie_qualitative}, showing fragmented word selections without 
structural coherence. Color coding follows Figure~\ref{fig:movie_qualitative}. 
Annotation precision scores indicate alignment with human rationales.}
  \Description{movie_qualitative_other}
  \label{fig:movie_qualitative_other}
\end{figure*}

\end{document}